\documentclass[journal]{IEEEtran}

\usepackage{graphicx}
\usepackage{multirow}
\usepackage{subcaption}
\usepackage{amsmath, amssymb}
\usepackage{cite, balance}
\usepackage{booktabs, xcolor}
\usepackage{stfloats, tabularx, caption}
\usepackage[inline]{enumitem}
\usepackage{float,xspace}
\usepackage[pagebackref,breaklinks,colorlinks]{hyperref}

\newcommand{\etal}{\textit{et al.}\ }
\newcommand{\ie}{\textit{i.e.,}\xspace}
\newcommand{\eg}{\textit{e.g.,}\xspace}
\newcommand{\vtext}[3]{
  \rotatebox{90}{%
    \colorbox{#2!20}{%
      \parbox[c][1.5em][c]{#1em}{\centering \textbf{#3}}
    }%
  }%
} %

\begin{document}

\title{Review of Demographic Fairness\\in Face Recognition}

\author{Ketan~Kotwal,~\IEEEmembership{Senior~Member,~IEEE,} and~S{\'e}bastien~Marcel,~\IEEEmembership{Fellow,~IEEE}
\thanks{Authors are with Idiap Research Institute, 1920-Switzerland. S.\ Marcel is
also with University of Lausanne, 1015-Switzerland.}%
\thanks{Manuscript received mm dd, 2025; revised mm dd, 2025.}%
}
%

\markboth{IEEE Transactions on Biometrics, Behavior and Identity Science}%
{Kotwal {\it and} Marcel: Review of Demographic Fairness in Face Recognition}

\maketitle
\begin{abstract}
The issue of difference in face recognition (FR) performance across demographic
groups has emerged as a critical area of research, given its impact on fairness,
equity, and reliability across diverse applications. As FR technologies are
increasingly deployed globally, disparities in performance across demographic
groups-- such as race, ethnicity, and gender-- have garnered significant
attention. These differences or biases not only compromise the credibility of FR
systems but also raise ethical concerns, especially when these technologies are
employed in sensitive domains. This review consolidates extensive research
efforts providing a comprehensive overview of the multifaceted aspects of
demographic fairness in FR.

We systematically examine the primary causes, datasets, assessment metrics, and
mitigation approaches associated with performance differences in FR across
demographic groups. By categorizing key contributions in these areas, this work
provides a structured approach to understanding and addressing the complexity of
this issue. Finally, we highlight current advancements and identify emerging
challenges that need further investigation. This article aims to provide
researchers with a unified perspective on the state-of-the-art while emphasizing
the critical need for equitable and trustworthy FR systems. 
%

\end{abstract}



\section{Introduction}
\label{sec:intro}

Differences in the performance of face recognition (FR) system  across
demographic groups, commonly described under the umbrella of demographic
fairness, have emerged as a critical challenge in the deployment of biometric
technologies for real-world applications~\cite{rathgeb2022demographic,
sixta2020fairface, drozdowski2020demographic, jain2021biometrics}.
Several real-world incidents underscore the societal risks associated
with such disparities. For example, a Black individual was wrongly arrested in
Detroit due to a false FR match~\cite{hill2020wrongfully}; a US civil liberties
test reported misidentification of a handful US lawmakers as criminals by major
FR service~\cite{aclu2018rekognition}; and recently, researchers exposed racial
bias in airport FR systems used during passenger boarding~\cite{marshall2023why,
falk2024struggle}. These systems frequently exhibit unequal recognition
performance across race, gender, and age groups~\cite{jain2021biometrics,
mehrabi2021survey, howard2019effect}. Such disparities, often labelled as
biases, can have far-reaching consequences, especially in critical applications
like border crossing, law enforcement~\cite{limante2024, jones2020law},
security~\cite{leslie2020understanding}, and hiring
processes~\cite{burgess2022watching, raji2020saving, pena2020bias}, where both
fairness and accuracy are paramount.

\begin{figure}[!t]
\centering
\begin{subfigure}{0.9\columnwidth}
\includegraphics[width=\columnwidth]{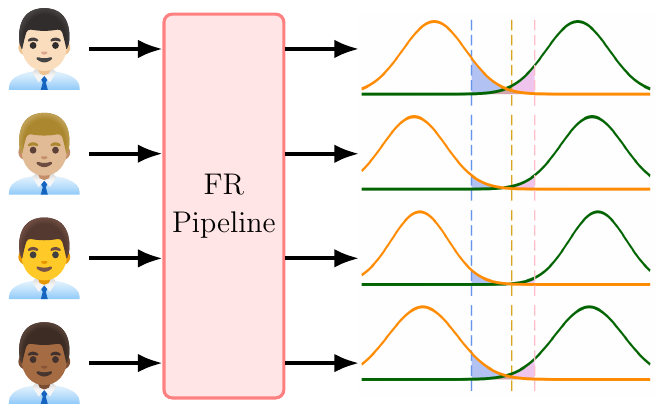}
\caption{}
\label{fig:race1}
\end{subfigure}
\begin{subfigure}{0.8\columnwidth}
\includegraphics[width=\columnwidth]{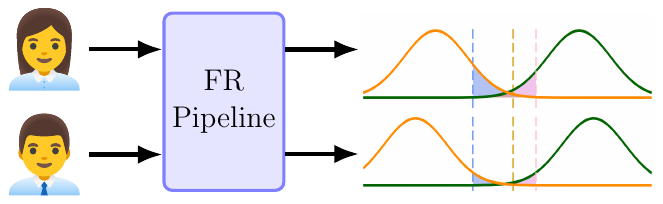}
\caption{}
\label{fig:gender1}
\end{subfigure}
\begin{subfigure}{0.8\columnwidth}
\includegraphics[width=\columnwidth]{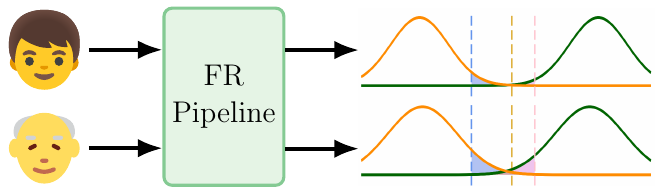}
\caption{}
\label{fig:age1}
\end{subfigure}
\caption{
Illustration of differences in performance across demographic groups in FR
with reference to different demographic factors: (a) race or ethnicity,
(b) gender, and (c) age. Score distributions for mated and non-mated
pairs are shown for multiple demographic groups, along with decision
thresholds (dotted lines) corresponding to different operating points.
This figure is intended to demonstrate that accuracy differentials may
persist across a range of thresholds and are not restricted to a
particular type of error (\eg false match or false non-match).}
\label{fig:intro}
\end{figure}

A conventional FR system evaluates the similarity between two facial images by
computing a matching score based on their extracted features to determine
whether they correspond to the same individual or not. The performance of such
systems is typically examined by analyzing the distributions of matching scores
for image pairs of the same identity and those of different identities. Scores
are thresholded into binarized decisions: match or no-match.
Fig.~\ref{fig:intro} illustrates the issue of demographic bias based on race,
gender, and age. Ideally, a fair FR model should exhibit equitable performance
across all demographic groups (\ie similar error rates or comparable score
distributions). A commonly implicit but rarely emphasized assumption is
that the groups being compared are equivalent in all aspects except for the
demographic attribute of interest. For instance, comparing a group of persons
wearing face masks with a group without masks- one may introduce additional
variability unrelated to demographics, making the comparison unsuitable for
assessing fairness. The disparity between \textit{equivalent} groups contributes
to the unfair or biased FR. (The formal definitions for error rates and
disparities provided in Sec.~\ref{sec:bias_intro} and~\ref{sec:assessment}.) As
depicted by various choices of scoring thresholds, accuracy differences exist
across demographic splits and can manifest at various thresholds. This
emphasizes that demographic disparities are not tied solely to false match or
false non-match errors but may be present throughout the score space. This issue
is further compounded by the growing reliance on FR technologies, making it
imperative to identify, evaluate, and mitigate sources of accuracy differences
effectively.

Due to its severity and the widespread range of FR applications, demographic
fairness has emerged as a crucial area of research, drawing significant
attention from both the biometrics and computer vision
communities~\cite{drozdowski2020demographic, jain2021biometrics, ross2019some,
busch2024challenges}. This issue has been formally incorporated into the
evaluation frameworks of prominent initiatives. The National Institute
of Standards and Technology (NIST) is a US agency that independently evaluates
biometric technologies. Its Face Recognition Vendor Tests (FRVT) benchmark the
accuracy of FR algorithms using standardized procedures and extensive datasets.
Since 2019, FRVT reports have incorporated analyses of demographic
disparities~\cite{frvt3, grother2021demographic}, making them a key reference
for assessing fairness in FR. Other initiatives, including the Maryland Test
Facility (MdTF) and the European Association for Biometrics (EAB), have also
contributed to this discourse, though at a smaller scale and scope compared to
NIST. The MdTF, supported by the United States Department of Homeland Security
(DHS), has conducted biometric technology rallies to evaluate demographic
disparities in FR systems. In Europe, organizations such as the EAB have hosted
dedicated events on demographic fairness in biometric systems, underlining the
global importance of this topic. This research area is often positioned within
the broader context of fair and trustworthy biometrics, and has been receiving
substantial attention-- in the form of papers, workshops, or special sessions--
from leading conferences such as IEEE/CVF CVPR, WACV, IEEE FG, ICPR,
ICLR, and ICML; and from reputable journals including IEEE Transactions
on Information Forensics and Security (TIFS), IEEE Transactions on Biometrics,
Behavior, and Identity Science (TBIOM)~\cite{9833480} and IEEE Signal Processing
Magazine~\cite{cheong2021hitchhiker}. Additionally, standards organizations have
recognized the need for systematic approaches. For example, the recently
published ISO/IEC 19795-10 guidelines quantify demographic differentials in
biometric systems, and emphasize the need for addressing fairness in FR
technologies~\cite{british2023iso}.\\

\noindent\textbf{Scope:} Fairness and bias in machine learning are expansive
topics, and their application in biometrics has drawn significant attention in
recent years. Several comprehensive reviews have addressed fairness and bias in
machine learning broadly~\cite{mehrabi2021survey, pessach2022review}, while
others focus specifically on biometrics, offering insights into various
modalities such as face, fingerprint, and vein, alongside applications beyond
recognition, including region of interest (ROI) detection, quality assessment,
and presentation attack detection\cite{drozdowski2020demographic,
jain2021biometrics, ross2019some, yucer2023racial}. However, as face remains the
most commonly used biometric trait in operational scenarios, a substantial
portion of research on differences across demographic performance or bias has
concentrated on this modality. 

While previous reviews offer broad perspectives, the extensive literature and
emerging challenges specific to demographic fairness in FR necessitate a
dedicated review. In this article, we provide a consolidated discussion of
recent advancements in the field, addressing the causes of demographic bias,
available datasets for research, evaluation metrics for assessment of fairness,
and recent mitigation techniques. Furthermore, we explore ongoing challenges
that persist in addressing demographic disparities, particularly in the light of
novel use cases and emerging FR applications. Although our primary focus is on
race and ethnicity, as these are dominant areas of research, we also include
gender-related studies within the broader context. Age-related studies,
given their distinct nature (variation over time for an individual) and
established body of research, are referenced only where directly relevant.\\

\noindent\textbf{Naming Conventions:} 
The primary objective of this review is to examine performance
differences in FR across demographic groups. Here, the term
\textit{performance} is used broadly to encompass any metric relevant to FR
evaluation, although it commonly refers to accuracy. The term
\textit{differences} denotes a quantifiable amount of the given performance
metric measured on a given dataset. In some instances, we use the term
\textit{demographic bias} to describe performance differences observed across
demographic groups (\eg race, gender, age), which may be reflected in metrics
such as accuracy or error rates. This terminology is used to align with common
usage in recent literature and is not meant to suggest deliberate discrimination
or convey any moral judgment.

Additionally, terms such as \textit{race} and \textit{ethnicity} are often used
interchangeably, although they represent distinct concepts. For clarity, we
retain the terms employed by the original studies. Similarly, the names of
ethnic groups vary across the literature: for instance, some works use terms
like Black and White, while others prefer African (or African-American) and
Caucasian. Additionally, the South Asian group is sometimes referred to as
Indian, whereas Asians often refer to East Asians. To maintain consistency and
respect the source material, we adhere to the original terminologies in this
review. Readers are encouraged to refer to the cited works for precise
definitions and context. While fairness in related topics such as face
detection, image quality, expression recognition, and attribute estimation is of
interest to the research community, this review exclusively focuses on
demographic fairness in FR.\\

\noindent\textbf{Contributions:} This review constitutes a comprehensive
work dedicated to exploring demographic fairness in FR, offering a unified and
holistic perspective to researchers in the field. We systematically analyze and
organize key aspects, including the causes of demographic disparities, available
datasets, assessment metrics, and mitigation techniques, providing a structured
framework for understanding these areas. Finally, we identify emerging
challenges and unresolved questions, inviting further research and innovation to
advance equitability in FR systems.

The structure of this paper is as follows: We begin with preliminaries of
assessment of generic FR in Section~\ref{sec:bias_intro}.
Section~\ref{sec:causes} explores the causes of differences in performance,
analyzing factors such as distribution of demographic groups in datasets,
skin-tone, image quality, and soft attributes. We provide an overview of
datasets commonly used for fairness-related research in
Section~\ref{sec:datasets}, highlighting their demographic attributes and
suitability for specific tasks. Section~\ref{sec:assessment} reviews existing
metrics for fairness evaluation, discussing their strengths and limitations.
Section~\ref{sec:mitigation} outlines recent bias mitigation strategies across
different stages of the FR pipeline. We discuss open challenges and future
research directions in Section~\ref{sec:future}, and conclude the review in
Section~\ref{sec:conc}.
%

\section{Preliminaries of FR Evaluation}
\label{sec:bias_intro}

Before delving into the fairness aspect of FR systems, we briefly outline the
standard FR pipeline and its associated evaluation terms. In a typical
verification setting, two face images are first processed through detection and
alignment stages to conform to the FR model's input requirements. The aligned
faces are then passed through a feature extractor, usually a deep neural
network, which produces a compact \textit{embedding}--a low-dimensional
representation of the subject's facial features. The similarity between two such
embeddings is computed (commonly using cosine similarity or Euclidean distance),
resulting in a continuous score typically bounded in [0, 1]. In many
applications, particularly those involving enrolled users (\eg access control),
the embeddings are precomputed and stored in a secure gallery. For
identification tasks, the query embedding is matched against a set of gallery
templates to determine the identity.

\begin{figure}[!t]
\centering
\includegraphics[width=0.9\columnwidth]{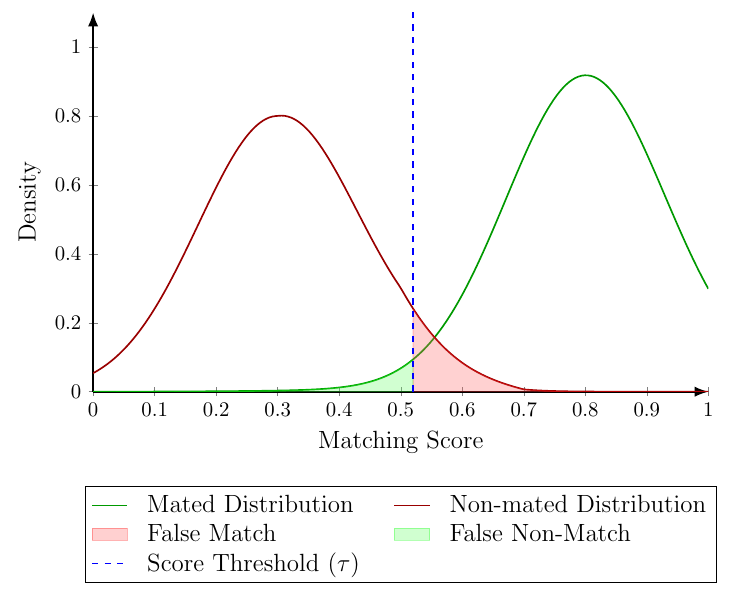}
\caption{Illustration of false matches and false non-matches arising from
distributions of mated and non-mated scores along with the score
threshold.}
\label{fig:score_dist_plot}
\end{figure}

In this context, a \textit{genuine} match (or mated pair) refers to two samples
originating from the same individual, while an \textit{impostor} match (or
non-mated pair) involves samples from different individuals. The similarity
score is binarized using a threshold (often denoted as $\tau$), producing a
match or non-match decision. The False Match Rate (FMR) is the proportion of
impostor pairs incorrectly classified as matches (false positives), while the
False Non-Match Rate (FNMR) is the proportion of genuine pairs incorrectly
classified as non-matches (false negatives). These two metrics are fundamental
to evaluating the accuracy and reliability of FR systems. Their values depend on
the chosen threshold $\tau$, with lower FMR and FNMR generally indicating better
performance. The overlap between the genuine and impostor score distributions
(see Fig.~\ref{fig:score_dist_plot}) defines the regions where these
classification errors arise.

The selection of the score threshold has a direct impact on the error rates in
face recognition. Increasing the threshold tends to reduce the False Match Rate
(FMR) by making it harder for impostor pairs to be incorrectly accepted as
matches, but this comes at the cost of a higher False Non-Match Rate (FNMR), as
more genuine pairs may be rejected, and vice versa. It is important, however, to
distinguish between the inherent overlap in the score distributions and the
binary classification decision that results from thresholding. The overlap in
the range of similarity scores between mated and non-mated pairs indicates
intrinsic ambiguity in the matching process, while the threshold simply
determines the point of separation between match and non-match decisions.

Two demographic groups may exhibit a similar degree of score overlap yet yield
different FMR and FNMR values due to differences in the shapes of their score
distributions or disparities in group-specific population characteristics.
Conversely, distributions across groups may differ significantly, while their
overall error rates remain similar. These variations-- in either score
distribution overlap or resulting error rates-- across demographic subgroups
form the foundation of what is commonly referred to as bias in FR systems. This
review focuses on analyzing, evaluating, and mitigating these disparities.
%
%

\section{Causes of Performance Differences}
\label{sec:causes}

In this section, we consolidate findings from existing works related to causes
of performance differences in FR. Considering the wide range of research in this
area, we have grouped the causes into categories for clarity and systematic
review. These categories encompass factors such as imbalances in training
datasets, variability in skin-tones, algorithmic sensitivity, image quality,
related covariates, combined or intersectional factors, and soft attributes.
Though this categorization simplifies the organization, it is important to note
that many studies attribute accuracy differences to multiple, overlapping
factors, making strict classification difficult. We have categorized works based
on their primary focus or findings, but some studies have been referenced across
multiple categories to reflect their broader relevance. This approach also
highlights that causes of demographic differences or \textit{bias} are
inherently multifaceted and interconnected, requiring a thorough understanding
to analyze and mitigate these disparities. A quick summary of various works
discussed in this section can be found in Table~\ref{tab:causes}.

\subsection{Training Datasets}

Early studies in this area speculated that the issue of performance differences
in FR systems across demographic groups often stems from imbalanced or
unrepresentative datasets, significantly influencing both training and
evaluation outcomes. Research by Krishnapriya \etal
\cite{krishnapriya2020issues} demonstrated how demographic groups, such as
African-American cohorts, exhibit higher FMRs, while Caucasian cohorts face
higher FNMR, highlighting the interplay between race and matching thresholds.
The FRVT conducted by NIST~\cite{frvt3} reported increased false positive
identifications in women, children, and the elderly, alongside higher false
negatives in under-represented racial groups-- particularly Black, East Asian,
and Native American individuals-- emphasizing the intricate interactions between
dataset characteristics and demographic attributes.

In a pre-deep learning era, studies like Klare \etal~\cite{klare2012face}
examined the impact of demographic balance in training datasets on FR
performance. While their results indicated that underrepresented groups-- such
as females and Black individuals-- benefited from increased representation, they
also found that balancing the dataset alone did not fully eliminate disparities
in recognition accuracy at test time. To further mitigate this, they proposed
training exclusive cohort-specific models and using dynamic model selection
during inference, highlighting that data balance might often be complemented
with architectural or procedural adjustments to meaningfully reduce accuracy
differences. Cavazos \etal \cite{cavazos2020accuracy} identified how dataset
complexity and identification thresholds contribute to racial bias, such as the
need for higher thresholds for East-Asian faces to achieve comparable false
acceptance rates (FARs). Gwilliam \etal \cite{gwilliam2021rethinking} challenged
the prevailing assumptions about the necessity of balanced datasets by
demonstrating that skewed distributions favoring African faces reduced racial
differentials more effectively than balanced datasets. In~\cite{albiero2020does,
albiero2020does_ijcb}, Albiero \etal demonstrated that balancing training data
for both genders (male and female) did not help reducing the corresponding
differential in FR accuracy. Wu and Bowyer~\cite{wu2023should} expanded this
discussion, emphasizing that mere balance in identities or number of images is
insufficient to address disparities, highlighting additional factors like
brightness and head pose during dataset assembly. These factors--often
referred to as soft attributes--include physical traits, behavioral patterns, or
material accessories associated with a person. They are descriptive and can be
categorized or classified~\cite{donida2022soft}. The soft-biometric attributes
tend to influence the performance of FR systems.

Other works delved into specific aspects of demographic balance in datasets.
Wang \etal~\cite{wang2019racial} observed that even race-balanced datasets
failed to eliminate accuracy differentials, hypothesizing that certain
ethnicities are inherently more challenging to recognize. Kolla and
Savadamuthu~\cite{kolla2023impact} highlighted the influence of facial quality
and racial feature gradations on model fairness. Focusing on inter-sectional
bias, Muthukumar \etal~\cite{muthukumar2018understanding} identified structural
facial features as significant contributors, particularly for dark-skinned
females, over attributes like skin-tone or hair length. Cook \etal
\cite{cook2019demographic} further analyzed the role of image acquisition
conditions, noting how factors such as skin reflectance and environmental
conditions disproportionately affect darker-skinned individuals, thus advocating
for standardized acquisition protocols to mitigate differences in FR
performance.

Although several studies acknowledge that significantly imbalanced training data
contribute to unfair or biased FR models and training with balanced datasets
reduces disparities to some extent, there is a consensus that these are neither
the sole causes nor complete solutions to the broader issue of demographic
fairness. Availability of large-scale training data, coupled with deeper models
with better learning capacity, has made it possible to investigate the aspect of
datasets in a more detailed manner. Recent findings have indicated that the role
of balancedness of training data is limited, and other factors that are inherent
to the individuals or their soft attributes may have a bigger role.
Additionally, balancing training data (especially through sub-sampling majority
groups) may involve discarding large amounts of high-quality data, thereby
reducing overall training volume and potentially decreasing recognition accuracy
while reducing the accuracy across constituent demographic groups.

\subsection{Variability in Skin-tone}

The influence of skin-tone on the performance of FR systems has been extensively
studied, revealing significant demographic disparities. A common method
to categorize skin-tone for FR studies is the Fitzpatrick Skin Type
Classification, which divides skin types into 6 categories (Type I to VI) based
on skin's reaction to ultraviolet (UV) exposure and general
pigmentation~\cite{fitzpatrick1988validity}. While widely used due to its
simplicity, it was originally designed for dermatology, not computer vision.
Alternative categorization schemes include the von Luschan's chromatic scale, a
36-point system based on observed skin color using standardized tiles, and the
Monk Skin-Tone (MST) Scale, a 10-tone scale designed to represent a broader and
more inclusive range of skin-tones~\cite{monk2023monk}. Each scheme varies in
granularity and applicability, and ongoing research continues to assess their
effectiveness.

In~\cite{muthukumar2018understanding}, Muthukumar \etal identified notable
under-performance in recognizing dark-skinned females compared to other
demographic group for commercial classifiers. Their analysis attributed these
disparities to structural features such as lips, eyes, and cheeks, in addition
to skin-tone itself. In another study, Krishnapriya
\etal~\cite{krishnapriya2020issues} analyzed FMR and FNMR across skin-tone
groups and observed error rate variations among demographic cohorts; however,
they did not find conclusive evidence linking darker skin-tones directly to
higher error rates. Buolamwini and Gebru \cite{buolamwini18a} employed the
Fitzpatrick skin classification system to evaluate commercial gender classifiers
and reported the lowest accuracy for darker-skinned females. While this study
evaluated gender classification rather than identity recognition, it led to
broader discussions around fairness in facial analysis systems, including those
used in biometric applications.

The Biometric Technology Rallies organized by MdTF have offered comprehensive
insights into the role of skin-related factors impacting FR performance. Their
2019 report~\cite{cook2019demographic} emphasized skin reflectance as a more
significant predictor of performance disparities than race. Skin reflectance
refers to the measurable amount of light reflected from the skin surface, which
affects how well facial features are captured in images; whereas skin-tone
refers to perceived skin color~\cite{cook2019demographic}. Using systematic
linear modeling, their study demonstrated that darker skin-tones were associated
with longer transaction (processing overall pipeline) times and lower accuracy
in biometric systems. The longer transaction times were primarily attributed to
difficulties in the face detection or image acquisition stage under suboptimal
lighting conditions. Lower skin reflectance can reduce contrast, making it
harder for detection algorithms to localize the face, thus increasing processing
time. This dependency was found to vary substantially across systems,
highlighting important role of acquisition methods in determining the extent of
performance differences. Lu \etal~\cite{lu2019experimental} provided a
quantitative assessment of performance variations across five skin-tone groups,
identifying light-skinned individuals as the easiest to verify and
darker-skinned individuals as the most challenging. However, ambiguities in
defining skin-tone categories complicate direct evaluations, highlighting the
need for standardized classification metrics.

While many studies report that individuals with lighter skin-tones tend to be
recognized more accurately than those with darker skin-tones, there is no
consistent consensus that skin-tone is the primary driver of differences in FR
performance across demographic groups. Moreover, it is important to note that
most analyses rely on apparent or perceived skin tone--estimated via brightness
or proxy classification schemes--rather than direct, objective measurement.

\begin{table*}[b]
\centering
\renewcommand{\arraystretch}{1.2}
\scriptsize
\caption{Summary of works delving into various causes of demographic fairness in
  face recognition. As several works have identified multiple causes of such
  performance differences, we have categorized the works based on their primary
  focus or inference. For details, readers are encouraged to refer to the source
  materials. The demographic factors of primary interest are denoted as
  \texttt{ET}: Ethnicity or race, \texttt{GN}: gender or sex, \texttt{AG}: age;
  whereas \texttt{+} indicates study of more attributes.}
\label{tab:causes}
\begin{tabularx}{\linewidth}{m{5mm} l l p{1.7cm} p {1.7cm}  p{8.5cm} }
\toprule 
~ & \textbf{Reference}            & \textbf{Year}  & \textbf{Dataset} & \textbf{Attribute} & \textbf{Summary}\\ 
\midrule
\multirow{10}{*}{\vtext{23.5}{green}{Training Datasets}} &
Klare \etal \cite{klare2012face} & 2012   & PCSO & ET, GN, AG & Advocated for balanced datasets and exclusive cohorts to improve face recognition performance.\\
 & Muthukumar \etal \cite{muthukumar2018understanding} & 2018 & PPB & GN, ST & Identified structural facial features as primary contributors to intersectional bias for dark-skinned females.\\
 & NIST FRVT \cite{frvt3}     & 2019   & Private & ET, GN, AG, + & Reported increased false positives in women, children, and elderly, and higher false negatives in underrepresented groups.\\
 & Wang \etal \cite{wang2019racial} & 2019  & RFW & ET & Observed that race-balanced datasets do not fully eliminate bias, suggesting inherent challenges in recognizing certain ethnicities.\\
 & Cook \etal \cite{cook2019demographic} & 2019 & Private & ET, GN, AG, +  & Analyzed image acquisition conditions, noting the impact of skin reflectance and environmental factors on darker-skinned individuals.\\
 & Krishnapriya \etal \cite{krishnapriya2020issues} & 2020  & MORPH & ET, ST & Demonstrated demographic disparities in FMR and FNMR, with African-American cohorts having higher FMR and Caucasian cohorts higher FNMR.\\
 & Cavazos \etal \cite{cavazos2020accuracy} & 2020 &  GBU & ET & Highlighted how dataset complexity and thresholds affect racial bias, requiring higher thresholds for East-Asian faces.\\
 & Gwilliam \etal \cite{gwilliam2021rethinking} & 2021  & BUPT, RFW & ET & Showed that skewed distributions favoring African faces can mitigate racial bias better than balanced datasets.\\
 & Wu and Bowyer \cite{wu2023should} & 2023  & DemogPairs, RFW, BFW, & ET, GN, + & Emphasized that balancing identities and images alone is insufficient, stressing brightness and head pose considerations.\\
 & Kolla and Savadamuthu \cite{kolla2023impact} & 2023 & RFW & ET  & Highlighted the influence of facial quality and racial feature gradations on fairness in face recognition models.\\
\midrule
\multirow{4}{*}{\vtext{8.5}{orange}{Skin-Tone}} &
Muthukumar \etal \cite{muthukumar2018understanding} & 2018 &  PPB & GN, ST & Identified structural facial features and skin tone as key factors for dark-skinned females' underperformance.\\
 & Cook \etal \cite{cook2019demographic}  & 2019 & Private & ET, GN, AG, +  & Highlighted skin reflectance as a major predictor of FR disparities and emphasized acquisition methods' role.\\
 & Lu \etal \cite{lu2019experimental}  & 2019 & IJB-B, IJB-C & AG, GN, ST, + & Quantified performance variations across skin tone groups, noting challenges with darker skin tones.\\
 & Krishnapriya \etal \cite{krishnapriya2020issues}    & 2020 & MORPH & ET, ST & Examined FMR and FNMR across skin tones but found no direct causation between darker skin tone and higher errors.\\
%
\bottomrule
\multicolumn{6}{>{\footnotesize\itshape}r}{Continued on the next page.}
\end{tabularx}
\end{table*}%
\begin{table*}[t]
\ContinuedFloat
\centering
\renewcommand{\arraystretch}{1.2}
\scriptsize
\caption{\footnotesize{\it (Continued)}}
\begin{tabularx}{\linewidth}{m{5mm} l l p{1.7cm} p {1.7cm}  p{8.5cm}}
\toprule 
~ & \textbf{Reference}            & \textbf{Year}  & \textbf{Dataset} & \textbf{Attribute} & \textbf{Summary}\\ 
\midrule
\multirow{7}{*}{\vtext{13.5}{green}{Algorithmic Factors}} &
Phillips \etal \cite{phillips2011other} & 2011 &  FRVT & ET & Identified the ``other-race effect," where algorithms performed better on their respective majority racial groups.\\
 & Klare \etal \cite{klare2012face}       & 2012 & PCSO & ET, GN, AG & Designed dynamic face matchers to work with models trained on exclusive cohorts.\\
 & Ricanek \etal \cite{ricanek2015review} & 2015 & ITWCC & AG & Observed inefficiency in choice of algorithms towards recognizing children's faces (due to structural changes).\\
 & Nagpal \etal \cite{nagpal2019deep}     & 2019 &  MORPH, RFW, CACD, + & ET, AG & Showed that deep learning models encode in-group biases, mirroring own-race and own-age human biases.\\   
 & Albiero \etal \cite{albiero2020analysis} & 2020 & AFD, MORPH, Notre Dame & GN & Found gender-based biases in score distributions, with lower accuracy for women across balanced datasets.\\
 & Dooley \etal \cite{dooley2023rethinking} & 2023 & CelebA, VGGFace2, RFW  & GN  & Demonstrated that network architecture and hyperparameters inherently influence fairness in FR; and bias can arise from model inductive biases.\\
\midrule
\multirow{8}{*}{\vtext{16.5}{orange}{Image Quality}} &
Lu \etal \cite{lu2019experimental} & 2019 & IJB-B, IJB-C & AG, GN, ST, + & Analyzed multiple covariates; noted lighter skin tones consistently outperformed medium-dark tones.\\
 & Krishnapriya \etal \cite{vangara2019characterizing} & 2019 & MORPH & ET &Showed improving image quality reduces performance gaps between African-American and Caucasian cohorts.\\
 & Cavazos \etal \cite{cavazos2020accuracy} & 2020 & GBU & ET & Highlighted dataset complexity and decision thresholds' impact on racial bias and accuracy.\\
 & Albiero \etal \cite{albiero2020analysis} & 2020 & AFD, MORPH, Notre Dame & GN & Linked gender-based disparities to score distributions and identified confounding factors like cosmetics.\\
 & MdTF \cite{cook_tbiom,cook2023demographic} & 2023 & Private & ET, GN, AG, + & Found lower skin reflectance correlated with reduced accuracy and higher transaction times.\\
 & Wu \etal \cite{wu2023face}   & 2023 & MORPH & ET, GN, + & Demonstrated how brightness inconsistencies increase FMRs and diminish similarity scores.\\
 & Bhatta \etal~\cite{bhatta2024impact} & 2024 &  MORPH & ET, GN & Studied impact of blur and resolution for 1:N identification.\\
 & Pangelinan \etal \cite{pangelinan2025lights} & 2025 & MORPH & ET, GN, + & Balancing brightness improves similarity scores for females.\\
\midrule
\multirow{9}{*}{\vtext{20}{green}{Intersectional Factors}} &
Ricanek \etal \cite{ricanek2015review} & 2015 &  ITWCC & AG & Discussed recognition challenges due to structural changes in children's facial features over time.\\
 & El Khiyari \etal \cite{el2016face}    & 2016 & MORPH & ET, GN, AG & Observed lower face verification accuracy in younger individuals, females, and Black racial groups.\\
 & Best-Rowden \etal \cite{best2017longitudinal} & 2017 & LEO\_LS, PCSO\_LS & AG, GN & Found that males generally have higher genuine scores, but their performance declines faster with age.\\
 & Vera-Rodriguez \etal \cite{vera2019facegenderid} & 2019 & VGGFace2 & GN & Highlighted gender as a covariate, with males consistently outperforming females across demographics.\\
 & Albiero \etal \cite{fr_age} & 2020 & MORPH & ET, AG & Observed lack of consistent relation between recognition accuracy and age.\\
 & FRVT report \cite{frvt3}    & 2021 &  Private & ET, GN, AG, + &  Noted elevated false positives for children and elderly, especially among Asian and American Indian groups.\\
 & Atzori \etal \cite{atzori2022explaining} & 2022 & VGGFace2, DiveFace  & ET, GN, + & Explanatory framework for  evaluating the influence of both protected and non-protected attributes.\\
 & Sarridis \etal \cite{sarridis2023towards}      & 2023 & RFW & ET, GN, AG & Reported high error rates for African females over 60 years, highlighting compounded biases.\\
 & Cook \etal \cite{cook2023demographic} & 2023 & Private & ET, GN, AG, + & Demonstrated that age and skin lightness significantly influence recognition scores, compounded by illumination.\\
\midrule
\multirow{7}{*}{\vtext{9.5}{orange}{Soft-Biometric Attr.}} &
Albiero \etal \cite{albiero2022face} & 2022 & MORPH, MFAD & Face region & Studies impact of facial regions on bias patterns.\\
 & Bhatta \etal \cite{bhatta2023gender} & 2023 & MORPH, MFAD & Hairstyle  & Noted that hairstyle balance reduces gender bias.\\
 & Pangelinan \etal \cite{pangelinan2024exploring} & 2024 & Various & Face regions  & Higher FMR for females despite controlling other factors.\\
 & Ozturk \etal \cite{ozturk2024can} & 2024 & MORPH & Facial hair & Study of facial hairs on impostor/genuine scores.\\
 & Wu \etal \cite{wu2024facial} & 2024 & MORPH, BA-test & Facial hair & Analysis of facial hairs (occlusions) for recognition accuracy.\\
 & Mamede \etal \cite{mamede2024fairness} & 2024 & RFW (occluded) & Occlusions & Reported that occlusions amplify racial bias.\\
 & Kurz \etal \cite{kurz2025illusion} & 2025 & MAAD-Face &  Facial Attributes & Gender gap vanishes with matched attributes.\\
\bottomrule
\end{tabularx}
\end{table*}%
%

\subsection{Algorithmic or Model Factors}

While majority of sources of performance differences in FR can be traced back to
data, several studies have identified how specific algorithm-related choices can
amplify or introduce demographic disparities. These algorithmic elements often
interact with other factors/ attributes, making causal attribution challenging.
In this section, we examine works where algorithm-level factors are prominently
discussed as contributors to observed bias. Phillips \etal
\cite{phillips2011other} identified the ``other-race effect,'' where algorithms
developed in Western and East Asian contexts demonstrated superior performance
for their respective majority racial groups.  The said effect refers to an
observation that humans recognize faces of their own race more accurately than
faces of other races. This disparity persisted even when datasets were balanced,
hypothesizing that underlying biases may occur due to algorithmic design and
training processes. However, it may be noted that most of these findings are
based on systems developed over a decade ago, and their implications may vary
for modern deep learning-based FR models.  Klare \etal~\cite{klare2012face}
observed recognition challenges for specific demographic groups, including
females, Black individuals, and younger cohorts. They reported improved
performance when models were trained exclusively on specific demographic
cohorts. This improvement was supported by the integration of dynamic face
matchers within the recognition pipeline--a modification that reflects both
algorithmic adaptation and the influence of training cohort composition.
Similarly, several recent works proposed modifying features or decision
thresholds based on group information (see Sec.~\ref{sec:mitigation}). While
such approaches might lead to improved accuracy for under-represented groups, it
requires reliable access to demographic labels. This introduces few drawbacks:
external labeling increases operational cost and may be prohibited under privacy
or anti-discrimination regulations (\eg GDPR). Alternatively, predicting
demographic attributes automatically introduces another layer of uncertainty,
where misclassification can directly impact system fairness and performance.

In~\cite{nagpal2019deep}, Nagpal \etal demonstrated that deep learning models
encode in-group biases, mirroring human tendencies such as own-race and own-age
effects. By analyzing activation maps, they showed that these biases were
ingrained within the feature representations of the models. Based on these
observations, they hypothesized that classification models encode faces from
different racial groups in distinct ways, indicating the presence of bias.
Although the study did not isolate algorithmic design as the sole cause, it
highlighted how internal model behavior may reflect or reinforce demographic
disparities. Dooley \etal systematically evaluated the impact of network
architectures and hyper-parameters on fairness in
FR~\cite{dooley2023rethinking}. Their findings revealed that architectural
choices significantly influence demographic disparities. It also demonstrated
that bias can be inherently encoded in a model's inductive bias, and relying
solely on standard architectures optimized for accuracy may compromise fairness.

Further investigations into gender-based disparities by Albiero \etal
\cite{albiero2020analysis} revealed skewed impostor and genuine score
distributions resulting in a lower accuracy in women. This disparity persisted
across datasets, regardless of balanced training and neutral facial expressions,
and hence, the authors speculated that the cause could be more intrinsic factors
between the groups. Ricanek \etal \cite{ricanek2015review} noted unique
challenges in recognizing children's faces due to structural changes with age,
finding that algorithms effective on adult faces performed poorly for younger
subjects. Although this issue pertains to longitudinal variations in facial
morphology, it underscores a broader implication: the same algorithms may not be
equally effective across seemingly similar tasks (FR). Attributes such as
facial structure should therefore be considered as critical factors in guiding
the design and selection of appropriate recognition algorithms.

Identifying algorithmic factors that contribute to differential recognition
accuracy across demographic groups remains a complex task. This process requires
isolating the influence of the training algorithm--an integral component of the
recognition pipeline. To ensure valid attribution, experimental setups need to
be carefully controlled that differ in nothing but algorithms. Initially,
performance disparities were primarily attributed to handful factors such as
imbalances in training datasets or variations in acquisition conditions.
However, as the number of potential influencing factors grows, the experimental
studies to rigorously investigate and isolate the algorithmic sources of
performance differences require a thorough and systematic approach.

\subsection{Image Quality}

The quality of input images and associated covariates significantly influence
the manifestation of demographic differentials in FR systems. Numerous studies
have emphasized how disparities in image quality across different demographic
groups can lead to variations in system performance. For instance, Cavazos \etal
\cite{cavazos2020accuracy} analyzed both data-driven-- focusing on aspects of
image quality-- and scenario-based factors, revealing that dataset complexity
and decision thresholds have a notable impact on recognition accuracy and racial
bias. Their experiments across multiple algorithms further demonstrated that
East-Asian faces required higher decision thresholds compared to Caucasian faces
to achieve equivalent error rates. The authors also acknowledged that image
quality may not act in isolation and that multiple underlying causes may jointly
contribute to observed disparities.

The study conducted by MdTF highlighted skin reflectance as a critical factor
influencing both the accuracy and efficiency of FR
systems~\cite{cook_tbiom,cook2023demographic}. Analyzing 158 FR systems, they
found that lower skin reflectance, typically associated with darker skin-tones,
correlated with reduced accuracy and higher transaction times. These effects
varied across systems, underscoring the role of image acquisition quality as a
stronger predictor of performance as mentioned earlier. Similarly, Wu \etal
\cite{wu2023face} explored the effects of brightness and illumination,
demonstrating that under-exposed or over-exposed images result in higher FMRs,
while significant brightness differences between image pairs diminish similarity
scores. They recommended controlled image acquisition processes to achieve
consistent brightness across demographic groups, thereby reducing accuracy
disparities. In~\cite{pangelinan2025lights}, authors analyzed the effect of
facial brightness on recognition performance disparities between Caucasian and
African American females. Their experiments showed that by balancing
brightness--using either median pixel values or full pixel distributions--the
similarity scores improve for both groups.

Krishnapriya \etal \cite{vangara2019characterizing} further examined how
variations in image quality contributed to performance gaps between
African-American and Caucasian cohorts. Enhancing image quality notably reduced
these disparities, particularly by minimizing low-similarity errors within the
genuine distribution. Following ICAO compliance guidelines, they evaluated
biometric sample quality to support these findings. Albiero \etal
\cite{albiero2020analysis} investigated gender-based disparities in FR systems,
linking these to differences in genuine and imposter score distributions. They
also identified confounding factors such as cosmetics and image pose. Despite
using neutral and balanced datasets, their study revealed that such measures
alone were insufficient to fully eliminate observed disparities. Bhatta
\etal~\cite{bhatta2024impact} explored 1-to-many identification performance
under degraded image conditions resembling surveillance-quality probes. The
findings showed that blur and low resolution significantly increased false
matches, with demographic disparities being more pronounced between genders than
racial groups, underscoring the influence of image quality on recognition
fairness. The investigation of demographic fairness in the context of 1-to-many
(1:$N$) face identification has been comparatively under-explored. Another
recent study in area \cite{pangelinan2024analyzing} observed that the accuracy
variation patterns in 1:$N$ identification are similar to those reported in 1:1
verification scenarios. Furthermore, their analysis highlighted that
degradations in image quality--particularly due to blur or reduced
resolution--tend to increase the false positive identification rate (FPIR).

Work by Lu \etal provided a detailed analysis of the influence of covariates on
FR performance, incorporating variables such as skin-tone, age, gender, pose,
facial hair, and occlusion across three datasets and five FR
systems~\cite{lu2019experimental}. Their findings highlighted that skin tone
significantly affects verification accuracy, with lighter skin tones
consistently outperforming medium-dark tones. However, they also emphasized the
challenges posed by ambiguities in skin-tone classification, advocating for more
precise methodologies for performance assessments. In alignment with earlier
studies, Lu \etal observed that male subjects generally achieved better
recognition accuracy than female subjects. They attributed this disparity to
factors such as occlusion caused by longer hair and alterations in facial
appearance due to makeup. These observations corroborate prior findings
indicating that facial makeup can negatively impact recognition
accuracy~\cite{dantcheva2012can, kotwal2019detection}. Collectively, these
studies underscore that demographic fairness in FR systems is intrinsically tied
to image quality and related covariates, necessitating focused efforts to
address these issues systematically.

\subsection{Combined or Intersectional Factors}

In the preceding sections, we examined the individual factors contributing to
demographic differentials in FR. This section shifts focus to studies that
investigate the combined effects of multiple demographic attributes, such as
age, race, and gender. Vera-Rodriguez \etal \cite{vera2019facegenderid}
emphasized the significance of gender as a covariate in FR, observing that males
consistently outperform females across various demographic groups. These
findings highlight the necessity of addressing combined demographic factors to
achieve equitable outcomes.

Age-related disparities in FR performance have been linked to structural
transformations in facial features over time, particularly among children.
Ricanek \etal \cite{ricanek2015review} observed increased complexity of child
aging compared to adults, attributing recognition challenges to changes in
facial bone structure and the proportions of facial components. Additionally,
Best-Rowden and Jain \cite{best2017longitudinal} reported nuanced patterns in
age-related recognition performance, noting that while males generally exhibit
higher genuine scores, their performance declines more rapidly with age compared
to females. These observations underline the intricate interplay of demographic
attributes in shaping performance of FR systems. The intersection of age, race,
and gender significantly amplifies accuracy differentials in FR systems. In
\cite{sarridis2023towards}, Sarridis \etal identified a disproportionately high
number of incorrect verifications (both false positives and false negative)for
African females over 60 years compared to Caucasians, illustrating the
compounded effects of intersecting demographic factors. Similarly, El Khiyari
\etal \cite{el2016face} demonstrated that face verification accuracy is notably
lower for younger individuals (aged 18--30), females, and certain racial groups
such as Black individuals, highlighting the challenges posed by such
intersections of demographic factors. Liu \etal~\cite{Liu_2022_CVPR} observed
that multiple factors such as ethnicity, pose, occlusion, and blur jointly
impact FR accuracy. 

Among the few studies addressing explanatory frameworks,
\cite{atzori2022explaining} introduced a methodology aimed at evaluating the
influence of both protected and non-protected attributes on FR performance.
Here, protected attributes refer to human characteristics legally safeguarded
from discrimination (\eg race or gender), whereas non-protected attributes
include modifiable traits such as makeup or hairstyle. Their analysis revealed
that fairness outcomes exhibit considerable variability across multi-attribute
groupings. Notably, patterns observed in single-attribute evaluations frequently
diminish or invert when extended to analyses involving multiple attributes.

Algorithmic evaluations further reinforce these findings. The FRVT report
\cite{frvt3} observed elevated false positives among children and the elderly,
particularly within Asian and American Indian groups. These disparities were
intensified in low-quality imaging conditions, with younger and older
demographics experiencing higher error rates. Cook \etal
\cite{cook2023demographic} extended this understanding by showing that
self-reported demographic factors like age and measured skin lightness
significantly impact recognition scores, often compounded by environmental
factors such as illumination. However, more recent analyses, such as Albiero
\etal~\cite{fr_age}, suggest that recognition accuracy may not consistently
improve or decline with age. Their findings indicate that face matchers do not
exhibit a simple age-performance trend, challenging assumptions from earlier
works.

The inconsistent findings indicate that age-related performance differences in
FR remains insufficiently understood and lacks a clear consensus. Overall these
studies highlight the complexity of mitigating intersecting biases within FR
systems, underscoring the importance of incorporating age-specific and
multi-factor considerations in both algorithm development and evaluation
protocols.

\subsection{Soft-Biometric Attributes or Non-Demographic Factors}

While prior sections have explored to demographic factors influencing
disparities in FR performance, recent research has highlighted the critical role
of non-demographic factors-- such as facial hair, hairstyle, and makeup-- in
shaping recognition outcomes. These factors, though not inherently demographic,
are deeply intertwined with social and cultural norms that vary across gender
and ethnicity. In the context of FR, these soft attributes function as partial
occlusions or lead to shifts in the underlying data distribution, thereby
introducing additional challenges. It is well-documented that occlusions can
significantly degrade recognition accuracy in biometric and generic vision
applications~\cite{zeng2021survey, boutros2021mfr, kotwal2024latent,
kortylewski2021compositional}. However, when these occlusions vary
systematically across groups, they contribute to unequal recognition outcomes
and further exacerbate existing disparities, mimicking demographic bias.

Recent studies have demonstrated that many of the observed demographic
fairness in FR may in fact be driven by these correlated non-demographic traits.
Through extensive experiments, Kurz \etal showed that the gender gap in
recognition accuracy vanishes when males and females share the same facial
attributes, indicating that social appearance conventions, rather than gender
itself, are a primary factor \cite{kurz2025illusion}. Pangelinan \etal similarly
found that controlling for facial pixel information across genders improved
non-matching errors for females, pointing to the role of hairstyles and other
aesthetic choices in accuracy differentials~\cite{pangelinan2024exploring}.
Multiple works have rigorously examined the impact of facial hair, showing that
variations in facial hair significantly affect both impostor and genuine score
distributions, and that these effects persist even with balanced training
datasets\cite{ozturk2024can,wu2024facial}. In females, factors such as
hairstyle, facial morphology, and overall facial dimensions result in a reduced
number of effective face pixels, which has been associated with elevated false
match rates (FMR) compared to males~\cite{albiero2021gendered}. Bhatta
\etal~\cite{bhatta2023gender} demonstrated that balancing for hairstyle-induced
occlusions in test data can nearly eliminate the gender gap. Additionally,
Albiero \etal illustrated that recognition performance can vary across different
facial regions, further suggesting that visual composition and occlusions,
rather than demographic identity alone, play a pivotal role toward
recognition~\cite{albiero2022face}. Complementing these findings, Mamede \etal
\cite{mamede2024fairness} proposed the Face Occlusion Impact Ratio (FOIR)- a
novel metric to measure the influence of occlusions on recognition performance
across demographic groups. Their results show that occlusions impact demographic
groups unequally, with African individuals disproportionately affected- thereby
amplifying existing biases.

These collective insights underscore the significant influence of
non-demographic but demographically correlated appearance factors in shaping
recognition performance. While these factors are not inherently biological, but
rather linked to the demographic groups through social and cultural appearance
norms and styles. Understanding and accounting for these nuanced sources of
variation is essential to mitigate differential accuracy and build fair FR
systems.
%

\section{Datasets for the Study of Demographic Accuracy Differences}
\label{sec:datasets}

In evaluating and mitigating demographic fairness in FR systems, the selection
of suitable datasets plays an important role. Although numerous FR datasets
exist, those specifically intended for differential performance and
fairness-related tasks must include demographic labels associated with each
subject or identity. The datasets designed for tasks such as race, gender,
ethnicity, or age estimation are particularly useful when they include
demographic labels, as such factors are critical for assessing fairness. For
certain tasks (related to estimation or classification of attributes), having a
single image per subject may suffice, as training and testing can be conducted
separately for each image. However, FR models, especially state-of-the-art
systems, benefit from having multiple images (variations) of each identity to
train more robust feature extractors. The testing phase or deployment has two
primary modes: one-to-one (1:1) verification and one-to-many (1:$N$)
identification. In 1:1 matching, a probe image is compared against one or more
enrolled images of a claimed identity. For identification (1:$N$ matching), the
probe is compared against a gallery containing reference images from multiple
identities to determine the closest match. Occasionally, the gallery may be
composed of multiple images per identity.
 
\begin{table*}[!h]
\renewcommand{\arraystretch}{1.4}
\centering
\caption{Commonly used datasets for tasks related to demographic accuracy
  differentials in face recognition.}
\label{tab:datasets}
\begin{tabular}{ l l l l l }
\toprule
\textbf{Dataset Name}  & \textbf{Year} & \textbf{\# Images / Subjects} & \textbf{Demographic Labels} & \textbf{Typical Purpose}\\ \midrule
MORPH-II~\cite{bingham2017morph}  & 2008/2016   & 55,000 / 13,000  & (Male, Female), (Black, White, Asian, Hispanic) & Train/ Test\\ 
MEDS-II~\cite{founds2011nist}   & 2011   & 1,300+ / 518    & (Male, Female), (Asian, Native American, Black, White)  & Test\\ 
Adience~\cite{eidinger2014age} & 2014 & 26,580 / 2,284 & (Male, Female) & Test\\
AFD (Curated)~\cite{kai_curation_method}\cite{xiong2018asian}    & 2018   & 91,000+  / 1,878   & (Male, Female), Asian  & Train\\ 
VGGFace2~\cite{cao2018vggface2}    & 2018   & 3.31M  / 9,131    & (Asian, Black, Indian, White)  & Train/ Test\\ 
DemogPairs~\cite{hupont2019demogpairs} &   2019  & 10,800 / 600  & (Male, Female), (Asian, Black, White) & Test\\
RFW~\cite{wang2019racial}   & 2019   & 24,000 pairs / -  & (Asian, African, Caucasian, Indian) & Test\\ 
KANFace~\cite{georgopoulos2020investigating} & 2020 & 40 K (+ 44 K videos) / 1,045 & (Male, Female)  & Test\\
BUPT-BalancedFace~\cite{wang2020mitigating}  & 2020  & 1.3M / 28,000   & (Asian, African, Caucasian, Indian) (7K per race)   & Train\\ 
BUPT-GlobalFace~\cite{wang2020mitigating}  & 2020  & 2M+ / 38,000   & (Asian, African, Caucasian, Indian)  & Train/ Test\\ 
DiveFace~\cite{morales2020sensitivenets}  & 2020   & 120,000 / 24,000   & (Male, Female), (Asian, African, European) & Train/ Test\\ 
BFW~\cite{robinson2023balancing} \cite{robinson2020face}  & 2020   & 20,000 / 800   & (Male, Female), (Asian, Black, Indian, White) & Test\\ 
MORPH-III~\cite{albiero2021gendered}   & 2021     & 127,000+ / 26,401  & (Male, Female), (Black, White) & Train/ Test\\ 
CASIA-Face-Africa~\cite{muhammad2021casia}    & 2021 & 38,500+ / 1183    & (Male, Female), African & Train/ Test\\
\bottomrule
\end{tabular}
\end{table*}
%

These requirements significantly reduce the availability of datasets suitable
for assessing demographic fairness, as most FR datasets do not provide adequate
demographic labels or have highly skewed distributions of subjects across
different demographic groups. It is important to note that several prominent
bias evaluation efforts, such as those led by NIST and the MdTF, rely on
proprietary or restricted-access datasets. While these organizations have
conducted large-scale, operationally relevant evaluations, their limited
availability to the wider research community poses challenges for
reproducibility, benchmarking, and comparative analysis. In this section, we
outline recent and commonly used publicly available datasets that are relevant
for assessing and addressing demographic fairness in FR systems. The datasets
presented below are arranged in chronological order of their release.

\begin{itemize}

\item \textbf{MORPH}~\cite{bingham2017morph}: The MORPH dataset is one of the
  largest facial image datasets available in several variants and versions.
  MORPH-II is the most commonly referred academic version, comprising more
  than 55,000 images from more than 13,000 subjects. Despite its usefulness,
  it should be noted that the dataset is highly skewed in terms of gender and
  ethnicity, with a significant over-representation of male subjects (more
  than 46,000 images) and a limited number of female subjects (approximately
  8,500 images). 

  The MORPH-III dataset has seen increased adoption in recent research
  investigating demographic variations in FR~\cite{albiero2021gendered}. For
  improved data quality, the dataset was curated to remove duplicates, twins,
  and mislabeled entries, resulting in a refined set comprising 127,000+
  images of nearly 26,000 subjects. This curated subset provides a cleaner and
  demographically structured resource for evaluating bias.  For both versions,
  the datasets consist of mugshot-style facial images acquired under
  controlled conditions, featuring neutral expressions, approximately frontal
  poses, and consistent illumination. However, it has been noted to have
  demographic imbalances, particularly under-representation of certain groups
  (\eg Asian, Indian), which may affect fairness studies.

\item \textbf{Multiple Encounter Dataset (MEDS-II)}~\cite{founds2011nist}: The
  MEDS-II dataset is an extension of the MEDS-I dataset and was created to
  assist with the NIST Multiple Biometric Evaluation. The dataset includes
  over 1,300 images of 518 subjects, with many subjects having only a single
  image, limiting its usefulness for verification tasks. The MEDS-II dataset
  is dominated by male subjects of White and Black ethnicities. Despite its
  limitations in demographic diversity, it remains a useful resource for
  testing FR systems in real-world scenarios, especially where multiple
  encounters of a subject are available. Additionally, the dataset is
  considerably smaller compared to current benchmarks.

\item \textbf{Adience}~\cite{eidinger2014age}: The Adience dataset
  consists of 26,580 images from 2,284 individuals captured in unconstrained (in
  the wild) conditions using typical consumer devices. The dataset is
  annotated with identity, gender, and one of eight age group labels. While
  primarily designed for age and gender classification tasks, the dataset's
  natural variation in pose, illumination, and demographics makes it valuable
  for FR research. Sourced from nearly 200 publicly available Flickr albums,
  this dataset has a higher representation of female subjects.

\item \textbf{AFD (Asian Faces Dataset - Curated)}~\cite{kai_curation_method}:
  The Asian Faces Dataset (AFD) was developed using images scraped from the web,
  with a focus on frontal face images~\cite{xiong2018asian}. The curated
  version, provided by Zhang \etal~\cite{kai_curation_method} includes over
  42,000 images of 911 males and 49,000 images of 967 females. A gender
  classifier was used to filter out mislabelled images, and duplicate or
  near-duplicate images were removed. This curated dataset is useful for
  studying gender and ethnic bias in FR systems, specifically for models
  focused on the Asian demographic.  

\item \textbf{VGGFace2}~\cite{cao2018vggface2}: The VGGFace2 dataset is a
  large-scale FR dataset containing over 3.31 million images of 9,131 subjects.
  This dataset was annotated for gender (Male, Female) and ethnicity (Asian,
  Black, Indian, White) labels by Idiap Research
  Institute\footnote{\href{https://gitlab.idiap.ch/bob/bob.bio.face/- /blob/
  master/src/bob/bio/face/database/vgg2.py}{Annotations for VGGFace2 Dataset}}
  making it useful for fairness related tasks. This dataset offers
  significant advantages due to a large number of identities with variations
  in age, pose, and lighting; VGGFace2 improves over the original with more
  diversity too. The dataset, however, is significantly imbalanced, with a
  disproportionate representation of White and male subjects.
  Although this dataset offers high intra-class variation with over 3
  million images, it contains only slightly more than 9,000 unique identities,
  which may limit its utility for evaluating false positive rates in
  large-scale evaluations. Note that, the original download source has been
  removed by the creators as of 2024.

\item \textbf{DemogPairs}~\cite{hupont2019demogpairs}: The DemogPairs dataset is
  a validation set containing 10.8K images, divided into six demographic folds:
  Asian females, Asian males, Black females, Black males, White females, and
  White males. The dataset was specifically designed to evaluate the demographic
  disparities in FR models, offering 58.3 million evaluation pairs, including
  cross-demographic, cross-gender, and cross-ethnicity pairs. The DemogPairs
  dataset was constructed with rigorous demographic annotation and is a useful
  resource for testing the generalization of FR systems across diverse
  demographic groups.

\item \textbf{Racial Faces in-the-Wild (RFW)}~\cite{wang2019racial}: The RFW
  dataset is a benchmarking dataset designed to study racial fairness in FR
  systems. It consists of four subsets (African, Asian, Caucasian, and Indian),
  each containing about 3,000 individuals and 6,000 image pairs (these pairs
  have been defined by its creators). The dataset is specifically used for face
  verification tasks and includes balanced pairs of genuine (mated) and imposter
  (non-mated) images. The RFW dataset has been widely adopted by the research
  community to evaluate and compare the performance of FR algorithms across
  different racial groups. However, given the overlap between the RFW dataset
  and the widely used training dataset, MS-Celeb-1M, it is important to consider
  that the evaluation set may have been \textit{seen} by the model during
  training. The creators of RFW have also provided a list of non-overlapping
  images to address this issue.

\item \textbf{KANFace}~\cite{georgopoulos2020investigating}: The KANFace
  dataset comprises around 40,000 still images and 44,000 video clips (totaling
  roughly 14.5 million frames) of 1,045 individuals, collected in unconstrained
  settings. On average, each subject is represented by 39 images and over 13,000
  video frames. The dataset includes manual annotations for identity, age,
  gender, and kinship, with demographic metadata sourced from Wikipedia and
  IMDb. The dataset consists of 586 male and 459 female subjects.

\item \textbf{BUPT-BalancedFace}~\cite{wang2020mitigating}: The
  BUPT-BalancedFace dataset was constructed to address demographic fairness by
  ensuring race balance across the dataset. It contains approximately 1.3
  million images from 28,000 celebrities, with a balanced distribution of 7,000
  identities per race. The dataset was selected from MS-Celeb-1M
  dataset~\cite{guo2016ms} through the FreeBase and Face++ APIs, although it has
  been noted that the labels may contain noise. Due to its size and balanced
  nature, the BUPT-BalancedFace dataset has become a popular resource for
  training and fine-tuning FR models while mitigating race-related biases. Two
  key points should be considered: first, the demographic labels, which are
  based on nationality or API, contain some noise. Second, the dataset overlaps
  with MS-Celeb-1M dataset, which restricts certain experiments when using
  models trained exclusively on MS-Celeb-1M dataset.

\item \textbf{BUPT-GlobalFace}~\cite{wang2020mitigating}: In addition to
  the BUPT-BalancedFace dataset, the BUPT-GlobalFace dataset is also widely used
  in fairness tasks. Unlike its balanced counterpart, BUPT-GlobalFace is
  racially imbalanced, containing over 2 million images from approximately
  38,000 celebrities, with an over-representation of certain racial groups (\eg
  Caucasian and Asian). Although it reflects real-world distribution more
  closely, the imbalance makes it less suitable for bias mitigation training and
  more appropriate for evaluating bias under unconstrained, skewed demographic
  settings.

\begin{figure*}[!ht]
\centering
\begin{subfigure}{0.98\columnwidth}
\includegraphics[width=\columnwidth]{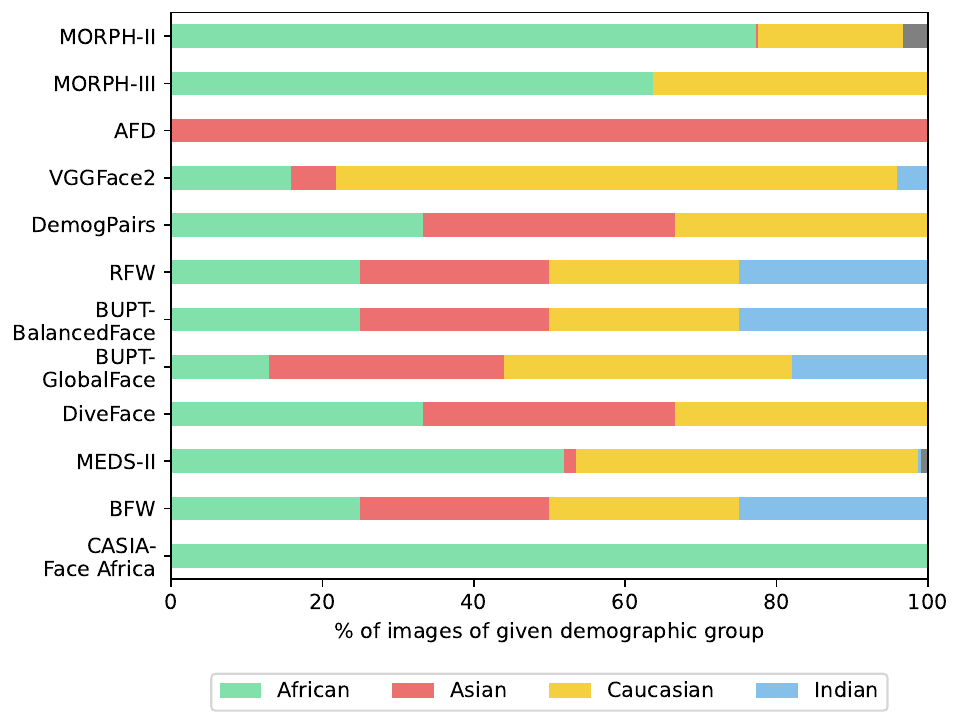}
\caption{}
\label{fig:race_dist}
\end{subfigure}
\hfill
\begin{subfigure}{0.98\columnwidth}
\includegraphics[width=\columnwidth]{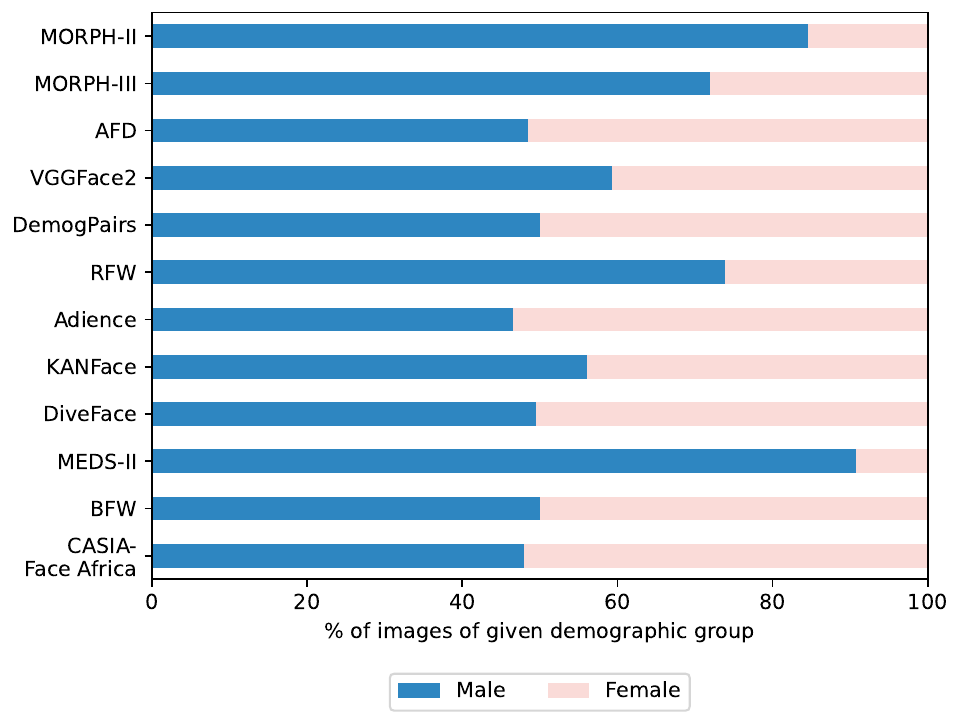}
\caption{}
\label{fig:gender_dist}
\end{subfigure}
\caption{Distribution of images of commonly used FR datasets considering (a)
  \textit{race} and (b) \textit{gender} as demographic factors. The details of
  distribution have been used from original sources (wherever available) or from
  other works contributing to this information; while the naming convention has
  been altered for unified representation aligning to the convention used by
  most datasets.}
\label{fig:dataset_dist}
\end{figure*}

\item \textbf{DiveFace}~\cite{morales2020sensitivenets}: The DiveFace is a
  dataset generated from the Megaface dataset (now
  decommissioned)~\cite{kemelmacher2016megaface}, containing over 120,000 images
  from 24,000 identities. The dataset includes two gender and three ethnicity
  classes, allowing for detailed demographic analysis. Annotations were made
  using a semi-automatic process, followed by manual inspection. This dataset is
  useful for studying the impact of gender and ethnicity in FR tasks, although
  it may exhibit bias in some groups. It should also be noted that in DiveFace
  dataset, the subjects of Indian and African ethnicities have been grouped
  together-- which can make it difficult to use it in conjunction with other
  datasets that typically do not follow such grouping.

\item \textbf{BFW (Balanced Faces in the Wild)}~\cite{robinson2023balancing}
  \cite{robinson2020face}: The BFW dataset was designed to provide a more
  balanced evaluation of FR systems by creating subgroups that are evenly split
  across gender and ethnicity. The dataset is compiled from VGGFace2
  \cite{cao2018vggface2} and offers a refined approach to subgroup analysis with
  less overlap between training and testing data. The corresponding demographic
  labels were generated using ethnicity~\cite{fu2014learning} and gender
  \cite{levi2015age} classifiers, followed by manual validation. Additionally,
  the BFW dataset is also balanced with respect to the number of images,
  subjects, and the (ratio of) images per subject; making it particularly useful
  for evaluating the demographic fairness of FR models, offering a more balanced
  alternative to other datasets.  However, this dataset shares the same two
  limitations as some earlier ones: a smaller size and overlap with VGGFace2.

\item \textbf{CASIA-Face-Africa}~\cite{muhammad2021casia}: The CASIA-Face-Africa
  dataset is the first large-scale face dataset of African subjects-- comprising
  38,546 images from 1,183 individuals, captured under varying illumination
  conditions using multi-spectral cameras. It includes detailed demographic
  attributes and facial expressions along with manually annotated with facial
  key points. The dataset exhibits a well-distributed age representation, with a
  significant portion belonging to the subjects up to 40 years, aligning with
  the majority workforce demographics. Additionally, it maintains an almost
  balanced gender ratio (48\% male, 52\% female), making it useful for
  gender-based analysis as well. In terms of ethnic variations, the dataset
  includes multiple African ethnic groups, with a notable dominance of the Hausa
  ethnic group. 

\end{itemize}

The datasets are summarized in Table~\ref{tab:datasets}, and
Fig.~\ref{fig:dataset_dist} illustrates an overview of the demographic
distribution, with race and gender as demographic variables (where details of
both variables were available), for the datasets reviewed in this section.
%
%

\section{Assessment of Demographic Fairness in FR}
\label{sec:assessment}

The bias in FR system leads to variations in score distributions and their
overlaps across different demographic groups. Such disparities inherently result
in different FMR and FNMR values for each group when a single (global) threshold
is used. Fig.~\ref{fig:intro} illustrates score distributions of different
groups may differ, leading to unequal error rates. In this section we briefly
review performance measures designed for demographic-aware assessments of
fairness in FR systems.
Given the overlap and distinct characteristics of scores and decisions,
establishing a well-defined evaluation framework is crucial for accurate
analysis. Howard \etal~\cite{howard2019effect} introduced the concepts of
differential performance and differential outcomes, providing two key terms that
aid in achieving a precise understanding and categorization of assessment
metrics.

\begin{itemize}
\item \textbf{Differential Performance:} Variations in genuine (mated) or
  imposter (non-mated) distributions across demographic groups, independent of
  thresholds.
\item \textbf{Differential Outcome:} Differences in FMRs or FNMRs between
  groups, based on decision thresholds.
\end{itemize}

Quantifying demographic bias or fairness, in terms of both demographic
performance and outcomes, is critical for developing fair and reliable FR
systems. By analyzing the overlap of distributions and error disparities, one
can identify specific areas requiring intervention to ensure equity across
diverse user groups.

Since the 2019 Face Recognition Vendor Test (FRVT)\footnote{Since 2023, the FRVT
initiative has been restructured into the Face Recognition Technology Evaluation
(FRTE) and Face Analysis Technology Evaluation (FATE) programs.} report, NIST
has included demographic effects in FR algorithms~\cite{frvt3}. This assessment
involved comparing non-mated pairs within the same demographic group, setting
thresholds for algorithms to achieve an FMR of 0.0001 for white males (since
this demographic typically associated with the lowest FMR).
Constraining non-mated pairs to share the same demographic attributes
(\eg race or gender) is commonly referred to as
\textit{yoking}~\cite{o2012demographic}. The report~\cite{frvt3} offered a
comprehensive analysis of recognition processes and identified areas where
demographic effects might occur. 
To quantify demographic disparity, NIST initially employed Inequity Ratios (IR),
calculating the ratio of maximum to minimum of FMR and FNMR, at score threshold
$\tau$, across demographic groups as shown in Eq.~\ref{eq:ir}.
\begin{align}
\text{IR}(\tau) = 
\left( \frac{\max_{d_i} \text{FMR}_{d_i} (\tau)}{\min_{d_j} \text{FMR}_{d_j} (\tau)} \right)^\alpha \times  
\left( \frac{\max_{d_i} \text{FNMR}_{d_i} (\tau) }{\min_{d_j} \text{FNMR}_{d_j}(\tau)} \right)^\beta,
\label{eq:ir}
\end{align}
--where, the subscripts $d_i, d_j$ refer to the demographic group; and $\alpha,
\beta$ are weighing coefficients. The error rates (FMR and FNMR) have
been \textit{yoked}, meaning they were calculated using pairs exclusively from
the same demographic group.

However, considering potentially large range of the error rates, these ratios
can become numerically unstable, especially in extreme cases. To alleviate this
shortcoming, NIST has considered few modified versions of Inequity Ratios such
as adjusting the score threshold ($\tau$), incorporating fixed constants in the
denominator, or expressing worst-case error rates relative to arithmetic or
geometric means~\cite{grother2022face}. Using the geometric mean is
particularly advantageous due to its extended range over FMR/FNMR values, and
this metric, as described in Eq~\ref{eq:ir_g} has often been used by NIST.
\begin{align}
&\text{A}(\tau) =  
\frac{\max_{d_i} \text{FMR}_{d_i} (\tau)}{\text{FMR}^{*} (\tau)} \nonumber \\
&\text{B}(\tau) = 
\frac{\max_{d_i} \text{FNMR}_{d_i} (\tau)}{\text{FNMR}^{*} (\tau)},
\label{eq:ir_g}
\end{align}
--where the error rates marked with $^*$ represent the geometric mean computed
across all demographic groups. In \cite{conti2024assessing}, the
authors computed the ratio of each group's error rate (FMR or FNMR) to the
geometric mean across all demographic groups. The overall bias metric was then
defined as the sum of the logarithms ($\log10$) of these ratios for all groups.
Another possible approach involves referencing a standard FMR or FNMR in the
denominator, which inherently resolves stability issues while providing more
robust evaluations of demographic fairness.

It is important to note that the IR metric defined in Eq.\ref{eq:ir}
combines two components: one derived solely from the group-wise FMR values and
the other from FNMR values. While the metric multiplies these components after
applying user-defined weights ($\alpha$, $\beta$), it is also possible to use
the individual terms independently (either or both). The constituent ratios (\eg
the max to min FMR or FNMR across demographic groups, in this case) provide
interpretable measures of disparity. The choice of weights typically depends on
the specific requirements of the end application. Several prior works have
evaluated fairness using single-component metrics focusing on either FMR or FNMR
separately\cite{frvt3, conti2022mitigating}. Most metrics, that assess
difference in FR performance, discussed below also derive from these two
fundamental error types. For brevity, we focus on combined formulations, but
note that using one-sided metrics (which is also equivalent to setting one
weight to minimum permissible value) is also a valid and reported practice when
appropriately motivated.

The Fairness Discrepancy Rate (FDR) is one of the initial efforts of quantifying
fairness in FR~\cite{de2021fairness}. For an FR system using a single decision
threshold, the FDR combines FMR and FNMR using weighted sums, and evaluates
fairness through a unified measure that captures the trade-offs between both
error rates. As described by Eq.~\ref{eq:fdr}, the FDR requires two
hyper-parameters: one for the score threshold ($\tau$), and another for defining
relative importance ($\alpha$) of FMR over FNMR. Thus, it offers flexibility of
assessing the fair nature of the model at pre-defined score threshold and
application-dependent weighing of false matches to false non-matches.
\begin{align}
&A(\tau) = \max \left | \text{FMR}_{d_i} (\tau) - \text{FMR}_{d_j} (\tau) \right |, \nonumber \\
&B(\tau) = \max \left | \text{FNMR}_{d_i} (\tau) - \text{FNMR}_{d_j} (\tau) \right |, \nonumber \\
&\text{FDR}(\tau) =  1 - \left(\alpha A(\tau)  + ( 1 - \alpha) B(\tau)\right).
\label{eq:fdr}
\end{align}

In a similar vein, to define the balance between false matches and false
non-matches, NIST proposed an approach to calculate the Inequity Measure by
raising the terms representing demographic disparities to specific exponents
(serving as weights) and then multiplying them. 

\begin{table*}[!h]
\renewcommand{\arraystretch}{1.4}
\centering
\caption{Summary of fairness assessment metrics employed in demographic analysis
  of face recognition systems.}
\label{tab:assessment}
\begin{tabular}{l l l p{7cm}}
\toprule
\textbf{Assessment Metric} & \textbf{Ref Publication} &  \textbf{Year} &\textbf{Description}\\ \midrule
Inequity Rate (IR) & NIST / Grother~\etal~\cite{grother2021demographic} & 2021 & This metric involves the analysis of the minimum and maximum ratios of False Match Rate (FMR) and False Non-Match Rate (FNMR), with the application of weighted exponents.\\ 
Fairness Discrepancy Rate (FDR) & Pereira \& Marcel~\cite{de2021fairness} & 2021 &  A weighted combination of the maximum differential values of FMR and FNMR.\\ 
Statistical Approaches & Schuckers \etal~\cite{schuckers2022statistical} & 2022 & Utilizes a bootstrap-based hypothesis testing approach to assess bias.\\ 
Separation/Compactness Metrics & Kotwal \& Marcel~\cite{kotwal2022fairness} &  2022 & Investigates the distributions of genuine and impostor scores, focusing on shape and compactness characteristics.\\ 
GARBE (Gini Coefficient Based Metric) & MdTF / Howard \etal~\cite{howard2022evaluating} & 2022 & Measures statistical dispersion in FMR and FNMR, integrating these through linear weighting.\\ 
Mean Absolute Percentage Error (MAPE) & Villalobos \etal~\cite{villalobos2022fair} & 2022 & Calculates the relative deviation of FMR from a pre-established benchmark.\\ 
Sum of Group Error Differences & Elobaid \etal~\cite{elobaid2024sum} & 2024 & Examines relative deviations in group-level FMR and FNMR compared to global scores, providing insight into disparities.\\ 
Comprehensive Equity Index (CEI) & Solano \etal~\cite{solano2024comprehensive} & 2024 & A weighted combination that assesses disparities in both tail and central distributions of performance metrics.\\ 
ROC Uncertainty  & Conti  \& Cl{\'e}men{\c{c}}on~\cite{conti2024assessing} & 2024 &  Utilizes normalized uncertainty of the ROC curves to assess fairness.\\
Standard Deviation & {\hfill - \hfill} &  - & The standard deviation of FMR, FNMR, and True Match Rate (TMR) is assessed to gauge variability.\\ 
d-prime ($d'$) & {\hfill - \hfill} & - &  Quantifies the separation between score distributions of different demographic groups; higher values indicate greater separability.\\
Skewed Error Ratio (SER) & {\hfill - \hfill} &  - & A ratio that compares the worst-case error rates across different demographic groups.\\ 
Mean Absolute Deviation (MAD) & {\hfill - \hfill} & - & Evaluates the MAD in specific error rate computed at a fixed threshold.\\
Trade-Off (TO) & {\hfill - \hfill} &  - & Evaluates the trade-off between performance metrics, particularly the difference in average accuracy and standard deviation.\\
\bottomrule
\end{tabular}
\end{table*}
%

%

Schuckers \etal highlighted the importance of accounting for statistical
variation when evaluating fairness in FR
systems~\cite{schuckers2022statistical}. They noted that the differences among
demographic groups can arise either from actual performance disparities or by
chance due to sampling variability, leading to potential Type-I errors. To
address this, they proposed two statistical methodologies: a bootstrap-based
hypothesis test and a simpler test methodology tailored for non-statistical
audiences. Their study also conducted simulations to explore the relationship
between margin of error and factors such as the number of subjects, attempts,
correlation between attempts, underlying FNMRs, and the number of demographic
groups.
In \cite{conti2024assessing}, the authors propose a statistical framework for
assessing fairness using Receiver Operating Characteristics (ROC) curves and
associated uncertainty measures derived from U-statistics. They argue that the
ROC curve is not only central to evaluating system performance but also serves
as a key tool for analyzing fairness in similarity scoring. They established
asymptotic guarantees for empirical ROC curves and fairness-relevant metrics,
emphasizing the importance of accurately quantifying uncertainty. Notably, they
highlight that conventional bootstrap methods may misrepresent uncertainty due
to the U-statistical nature of FMR and FNMR, and recommend a recentering
technique to ensure valid inference.
In~\cite{howard2022evaluating}, the researchers at MdTF proposed new metric for
assessment of bias based on demographic outcomes. Their metric, GARBE (Gini
Aggregation Rate for Biometric Equitability), is inspired by the Gini
coefficient-- which has a long history of use as a dispersion measure in
socio-economic context. The GARBE evaluates statistical dispersion in error
rates and emphasizes equitable treatment across demographic groups. Similar to
the FDR and IR, this metric combines weighted contributions of FMR and FNMR to
produce a single fairness score for a given fixed score threshold. The formula
for GABRE and its constituent Gini coefficient are provided in
Eq.~\ref{eq:garbe}.
\begin{align}
\text{GARBE} &= \alpha \, G_\text{FMR}(\tau) + (1 - \alpha) \, G_\text{FNMR}(\tau), \quad \text{where}, \nonumber\\
G_x &= \left( \frac{n}{n-1}\right) \left(\frac{\sum_{i=1}^{n} \sum_{j=1}^{n} | x_i - x_j |}{2 n^{2} \bar{x}} \right).
\label{eq:garbe}
\end{align}

Metrics such as GARBE and several others mentioned earlier provide a
single scalar value to quantify fairness. This reporting has many advantages as
it facilitates simplified, easy comparisons and ranking of systems. However,
reducing multiple fairness indicators into a single combined measure as well
introduces certain limitations. In particular, the choice of weighting
coefficients is often subjective or application-dependent, and may not reflect
the actual impact of each component. Moreover, combining ratios (\eg from FMR
and FNMR) that operate on different scales of magnitude can inadvertently skew
the metric toward one aspect of fairness, thereby distorting the overall
assessment. If the weights are not thoughtfully chosen and justified, such
aggregation may obscure significant disparities between demographic groups.

Villalobos \etal proposed the Mean Absolute Percentage Error (MAPE) as a metric
to quantify differences in error rates across demographic
groups~\cite{villalobos2022fair}. MAPE measures the relative deviation of FMRs
from a policy-defined FMR, ensuring that low error rates for one group do not
mask higher error rates for another. High deviations in error rates,
particularly towards lower values of FMR, can negatively impact the system by
increasing FNMR. A MAPE score of zero indicates that all demographic groups
achieve the desired FMR, making it an effective metric for fairness evaluation.
If $\text{FMR}_{p}$ indicates policy-defined FMR, then MAPE for a system using
$N$ groups can be computed as: 
\begin{equation}
\text{MAPE} = \frac{100}{N} \, \sum_{d=1}^{N} 
\left| \frac{\text{FMR}_d - \text{FMR}_{p}}{\text{FMR}_{p}} \right|.
\label{eq:mape}
\end{equation}

The Sum of Group Error Differences (SED\textsubscript{G}) was introduced as the
fairness assessment metric address disparities in biometric verification systems
in \cite{elobaid2024sum}. The SED\textsubscript{G} calculates relative
deviations in FMR and FNMR across demographic groups from the FMR/ FNMR of
global scores. They consider the Equal Error Rate (EER) threshold as a reference
to compute the error rates. In other words, it adapts a relative difference
formula to quantify demographic fairness by comparing individual group performances
to a global standard. Authors argue that by incorporating both
within-demographic (WDI) and cross-demographic (CDI) interactions,
SED\textsubscript{G} is able to provide better understanding of the
magnitude and type of bias making it a versatile measure. For a setup comprising
$\mathcal{D}$ set of demographic groups, the SED\textsubscript{G} can be
obtained as:
\begin{align}
&\delta X_d =  \left| 1 - \frac{X_d (\tau)}{X_\text{global} (\tau)} \right|, \cdots \quad X \in (\text{FMR}, \text{FNMR}), \nonumber \\
&\text{SED}_{\rm G} = \{ \delta \text{FMR}_d + \delta \text{FNMR}_d \,\, | \, d \in \mathcal{D} \}.
\label{eq:sedg}
\end{align}

Relatively fewer attempts have been made to assess the demographic fairness at
score-level (\ie based on differential performance). Among the various
metrics used to quantify fairness in biometric systems, d-prime ($d'$) has seen
increasing adoption~\cite{bhatta2024impact, bhatta2023gender, wu2023face}.
Originally from signal detection theory, it measures the distance between score
distributions of mated and non-mated pairs, and can be adapted to assess
separability between demographic group distributions. It is
threshold-independent, making it robust in cases where system decision
thresholds are not aligned across deployments. If $\mu, \sigma$ represent the
mean and standard deviations of the scores, the $d'$ is obtained as:
\begin{equation}
d' = \frac{|\mu_m - \mu_{nm}|}{\sqrt{\frac{1}{2}(\sigma_{m}^2 + \sigma_{nm}^2)}},
\label{eq:dprime}
\end{equation}
---where the subscripts $m$ and $nm$ refer to the distributions of mated and
non-mated pairs, respectively. The Kolmogorov--Smirnov (KS) bias metric
has been adapted for assessing demographic bias in FR by measuring disparities
between the score distributions of different demographic
groups~\cite{salvador2021faircal}.

In~\cite{kotwal2022fairness}, Kotwal and Marcel introduced three fairness
evaluation measures that emphasize the separation, compactness, and distribution
of genuine and impostor scores. Unlike conventional approaches that depend on
system accuracy, these measures focus on assessing differential performance
without requiring external parameters such as score thresholds. By examining how
\textit{well} the match is, rather than merely determining a match, this
approach provides a more nuanced evaluation of demographic fairness. The
Separation Fairness Index (SFI) measures the consistency in separation between
the mean genuine and impostor scores across groups--higher consistency implies
better fairness. The Compactness Fairness Index (CFI) assesses the similarity in
intra-group score variability by comparing the standard deviations of genuine
and impostor scores across demographics. The Distribution Fairness Index (DFI)
evaluates how closely the entire score distribution of each group matches a
reference distribution (\eg the population average), using a statistical
divergence measure. Their normal variants are summarized in Eq.~\ref{eq:fim}.

\begin{align}
\scriptstyle
& S(d) = | \mu_m (d) - \mu_{nm}(d)|; \text{and } C(d) = \sigma_m (d) + \sigma_{nm}(d);\nonumber \\
&\text{SFI} = 1 - \frac{2}{N} \sum_{d=1}^{N} \left| S(d) - \bar{S} \right|; \quad
\text{CFI} = 1 - \frac{2}{N} \sum_{d=1}^{N} | C(d) - \bar{C} | \nonumber \\
&\text{DFI} = 1 - \frac{1}{N \, \log_2 N} \sum_{d=1}^{N} \mathcal{D}_{\rm KL}\left( \mathbb{P}(d) \, || \, \mathbb{P}_{\text{ref}} \right)
\label{eq:fim}
\end{align}

Here, $N$ denotes the number of demographic groups; $\mu_m(d)$ and $\mu_{nm}(d)$
are the mean mated and non-mated scores for group $d$; $\sigma_m(d)$ and
$\sigma_{nm}(d)$ are their standard deviations; $\mathbb{P}(d)$ is the score
distribution for group $d$; $\mathbb{P}_{\text{ref}}$ is the reference
distribution (often the average across groups); and $\mathcal{D}_{\rm KL}$
represents the KL divergence.
Building on the work from \cite{kotwal2022fairness}, Solano \etal developed the
Comprehensive Equity Index (CEI)-- combining error rate differences and
recognition score distribution disparities~\cite{solano2024comprehensive}. The
CEI enhances bias quantification methods by considering both the distribution
tails and overall shapes of score distributions, enabling the detection of
subtle biases across demographic groups. They also conducted experiments on high
performing FR systems (as per NIST evaluations) using challenging, real-world
datasets.  Their experiments showed that CEI was able to effectively capture the
demographic bias on several challenging datasets with several covariates. In
compact manner, the CEI can be described as: $ \text{CEI} = w_1 \cdot \text{Tail
Disparity} + w_2 \cdot \text{Central Disparity} $. The disparity calculation is
similar to that of DFI expressed in Eq.~\ref{eq:fim}.

Several studies have evaluated bias and fairness in FR systems using the
standard deviation of performance metrics calculated across demographic groups.
These metrics include FMR, FNMR, and True Match Rate (TMR)-- where higher value
in standard deviation corresponds to greater demographic
disparities~\cite{terhorst2021comprehensive, gong2020jointly,
lu2019experimental, robinson2020face, wang2020mitigating, villalobos2022fair}.
The mean absolute deviation (MAD) of a metric across demographic groups
is another approach used to quantify bias. For example,
in~\cite{terhorst2020comparison}, Terh{\"o}rst \etal evaluated bias by computing
the MAD of the TMR at a fixed FMR threshold. Another significant metric is
the Skewed Error Ratio (SER), which specifically focuses on worst-case error
ratios, providing insights into the performance imbalance across
groups~\cite{villalobos2022fair, wang2020mitigating, kotwal2024WACV}. Recent
competitions~\cite{melzi2024frcsyn, melzi2024frcsyn1, deandres2024frcsyn}
exploring the use of synthetic data for FR and bias mitigation have adopted a
trade-off performance metric: the mean accuracy adjusted by the standard
deviation. This metric aims to ensure that efforts to mitigate bias do not come
at the expense of recognition performance. This metric emphasizes the
development of FR models that achieve both high recognition performance and
fairness across constituent demographic groups.

Table~\ref{tab:assessment} provides a brief summary of the fairness assessment
metrics discussed in this section. While several metrics listed here capture
similar goals: quantifying inter-group performance disparities, they differ in
sensitivity, interpretability, and need for threshold selection. Some, that
employ differential FMR/FNMR, are easier to relate to operational performance
but may not generalize across systems with differing threshold settings. Others,
like d-prime and fairness index measures, focus on distributional shifts and are
less sensitive to specific thresholds. An important yet often
under-addressed aspect in fairness evaluation is the choice of threshold setting
in threshold-based metrics. Most studies adopt a global threshold determined
from the overall population FMR or FNMR. Alternatively, some works use yoking,
where thresholds are computed using impostor pairs from the same demographic
group, aiming to reflect intra-group calibration. In certain cases, thresholds
are derived based solely on the FMR of a dominant group (\eg White Caucasians as
seen in FRVT~\cite{frvt3}). Some works determine the threshold by fixing the
maximum intra-group FMR to enhance robustness against demographic
imbalance~\cite{conti2022mitigating}. Since the threshold setting significantly
affects all threshold-based fairness metrics, it is a crucial design choice that
must be made with transparency and contextual awareness. As accuracy
differentials become narrower with improvements in model architectures and
training data, future fairness evaluations should include both threshold-based
and threshold-agnostic metrics.
%
%

\section{Bias Mitigation in Face Recognition Systems}
\label{sec:mitigation}

\begin{figure*}[!t]
\centering
\includegraphics[width=0.95\textwidth]{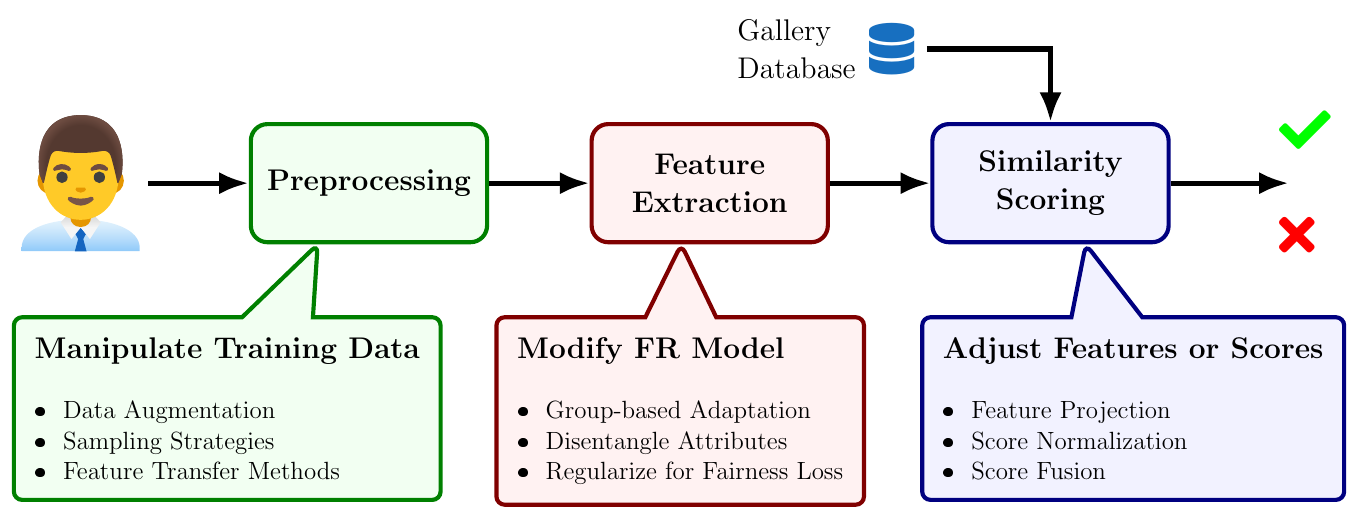}
\caption{Illustration of categories of bias mitigation methods in FR.}
\label{fig:mitigation_methods}
\end{figure*}
    
In an FR system, bias mitigation can be applied at different stages of the
recognition pipeline, providing a structured approach to addressing demographic
disparities. These stages align with general categories of bias mitigation
strategies commonly used in machine learning: pre-processing, in-processing,
and post-processing~\cite{pessach2022review, mehrabi2021survey,
singh2022anatomizing, hort2024bias}.

Pre-processing methods aim to address fairness at the level of training data by
modifying or augmenting it to reduce discriminatory patterns or imbalances
before the data is used for training. This approach is particularly relevant
when representational disparities in the data contribute to variations in FR
performance across demographic groups. The in-processing methods focus on
modifications to the FR model during training or fine-tuning phase, often by
incorporating constraints or objectives that optimize fairness without
significantly compromising recognition accuracy. Finally, post-processing
techniques involve adjustments to the output of trained models to ensure
fairness across demographic groups. These techniques modify the results, such as
classification scores or decision thresholds, to achieve equitable performance
without altering the underlying model. Figure~\ref{fig:mitigation_methods}
summarizes the categorization of bias mitigation methods-- that correlates with
the corresponding stages of FR pipeline. These methods are generally aimed at
minimizing accuracy disparities across demographic groups while maintaining
overall recognition performance. However, empirical evidence from most studies
indicates that efforts to reduce such disparities often lead to a reduction in
overall accuracy. Moreover, accuracy is often not the most meaningful metric in
deployment settings, where systems are evaluated at specific operating points
defined by fixing a particular error rate, such as FMR, FNMR, or TMR. As a
result, bias mitigation techniques should as well consider application-relevant
metrics evaluated at the corresponding operational thresholds.

\subsection{Pre-Processing Methods}

Pre-processing methods, also known as data-based methods, focus on modifying the
biometric samples before feeding into the FR system. These techniques aim to
normalize the characteristics of the data to make them robust for subsequent
feature extraction. Most of the methods in this category can also be regarded as
special case of data augmentation, specifically designed to reduce demographic
disparities in the training data.

In \cite{klare2012face}, Klare \etal showed that FR algorithms were performing
worse for female, Black, and younger individuals. To address this concern, they
proposed two mitigation strategies based on selection of training data. First,
training models on specific demographic cohorts to enhance recognition for those
groups, and second, implementing a dynamic face matcher selection approach,
where different algorithms, each trained on distinct demographic groups, would
be chosen based on probe information. A similar approach was followed by Deb
\etal in the context of longitudinal study of FR~\cite{deb2018longitudinal}.
Using the data of more than 900 children captured over time, they demonstrated
that recognition accuracy decreases as the time gap between image captures
grows. To address this, they fine-tuned FaceNet~\cite{schroff2015facenet} on a
separate child face dataset, showing improved accuracy. The authors advocated
the need for tailored training and evaluation of FR systems for different age
groups. Lu \etal \cite{lu2019experimental} investigated the influence of
covariates--such as age, gender, pose, and skin tone-- toward the performance of
face verification.  Their findings showed that using gender information to
curate training data improves performance, particularly at low FMRs.

In \cite{kortylewski2019analyzing}, Kortylewski \etal demonstrated the
effectiveness of using synthetic face images to mitigate the accuracy
differentials arising from real-world datasets. By utilizing synthetic data for
pre-training the FR model, they showed that the negative effects of dataset
bias, particularly with regard to pose variations, could be significantly
reduced. Their experiments revealed that pre-training the CNNs with synthetic
data, the need for real-world data can be reduced by up to 75\%. Although this
study did not directly address demographic fairness, we include it in the review
as it highlights the potential of synthetic data in improving the generalization
performance of FR systems and reducing (non-demographic) dataset bias. Yucer
\etal \cite{yucer2020exploring} proposed a GAN-based adversarial image-to-image
augmentation method to generate racially balanced training data at subject level
while preserving identity features. Evaluated across multiple loss functions,
their approach reduced performance variance across racial groups and improved
recognition accuracy for minority subjects.
Recently, Kotwal and Marcel proposed an Image-to-Image transformation module
called Demographic Fairness Transformer (DeFT), which enhances image
representations before passing them to pretrained CNNs
\cite{kotwal2024demographic}. The DeFT uses multi-head encoders and
soft-attention mechanisms to selectively enhance images based on inferred
demographic information. The demographic labels of race or ethnicity are often
non-discrete-- but this concern has rarely been addressed. In
\cite{kotwal2024demographic}, they replaced hard labels with probabilistic
weights which are implicitly inferred at run-time. Their experiments show that
DeFT reduces variations in FR performance across demographic groups and, thus
improves model fairness, with some models also achieving slightly better
accuracy compared to baseline systems.

\subsection{In-Processing Methods}

This category of works, also known as model-based methods, are applied during
the feature extraction stage by modifying the weights of the FR model. The goal
is to learn weights that generate features or embeddings that are less sensitive
to demographic differences.

Amini \etal \cite{amini2019uncovering} proposed a debiasing algorithm that
adjusts the sampling probabilities of data points in large datasets to reduce
hidden biases. In the face detection use-case, their algorithm led to a decrease
in race and gender bias while improving classification accuracy. To our
knowledge, similar approaches have not been tested for recognition or
verification applications.
In \cite{wang2020mitigating}, Wang and Deng introduced a reinforcement
learning-based race balance network (RL-RBN) where they applied adaptive margins
through deep Q-learning. Their method aimed to reduce the skewness of feature
scatter between racial groups, leading to more balanced performance across
different demographics. As a part of this work, they also released two
datasets--- BUPT-GlobalFace and BUPT-BalancedFace datasets--- that were
specifically designed to study racial bias in FR systems. 

Wang \etal explored the issue of highly-skewed class distributions in FR
datasets~\cite{wang2019deep}. They proposed Large Margin Feature Augmentation
(LMFA) and Transferable Domain Normalization (TDN) as methods to balance class
distributions by augmenting and normalizing the feature space. These methods
were shown to enhance the performance of underlying FR models by mitigating
issues arising from class imbalance, which often correlates with demographic
bias in under-represented groups.
In \cite{yin2019feature}, Yin \etal introduced a center-based feature transfer
framework to address the under-representation of certain demographic groups in
FR datasets. By transferring feature distributions from well-represented classes
to under-represented ones, they augmented the feature space for these groups,
reducing bias and improving recognition performance for under-represented
subjects. 

Gong \etal proposed a group-adaptive classifier (GAC) that uses adaptive
convolution kernels and attention mechanisms tailored to different demographic
groups \cite{gong2021mitigating}. By applying kernel masks and attention maps
specific to each group, their method activates facial regions that are more
discriminative for each demographic, thereby improving recognition accuracy and
fairness across demographic groups. 
Another approach by the same authors \cite{gong2020jointly} introduced an
adversarial network, called DebFace, which utilizes a multi-task learning
framework to simultaneously learn identity and demographic attributes. Their
method employed adversarial training to disentangle identity features from
demographic attributes such as gender, age, and race to effectively reducing
differential in the recognition process. Their experiments demonstrated that,
for DebFace, not only recognition but also demographic attribute estimation
tasks were less biased.
Huang \etal \cite{huang2023gradient} introduced GABN, a de-biasing
network that combines gradient attention maps (GAM) with adversarial learning to
mitigate racial bias in face recognition. The method enforces consistency of
GAMs across racial groups and uses GAM-guided sensitive region erasure to
enhance feature learning for darker-skinned individuals. Their experiments
showed that GABN was effective at reducing the performance gap between
darker-skinned subjects and Caucasians.
In~\cite{Ma_2023_ICCV}, authors proposed unsupervised data partitioning, in
iterative manner, followed by Invariant Feature Regularization, to improve
generalization across demographic groups. The unsupervised partitioning was
chosen as an inexpensive alternative to annotating training data with multiple
(confounding) attributes.

\begin{table*}[!b]
\renewcommand*{\arraystretch}{1.2}
\centering
\caption{Summary of recent works on mitigation of demographic bias in face
  recognition.}
\label{tab:mitigation}
{\small
\begin{tabularx}{\linewidth}{l l l p{2cm} p{8cm}}
\toprule
\textbf{Reference}    & \textbf{Year} & \textbf{Type of Method} & \textbf{Test Dataset} & \textbf{Summary}\\ \midrule
Klare \etal \cite{klare2012face}                 & 2012    & Data-Processing   & PCSO   & Training models on specific demographic cohorts\\
Deb \etal \cite{deb2018longitudinal}             & 2018    & Data-Processing   & CLF     & Finetuning models on specific cohort (age, in this case)\\
Lu \etal \cite{lu2019experimental}               & 2019    & Data-Processing   & IJB-B, IJB-C    & Curation of training data\\
Kortylewski \etal\cite{kortylewski2019analyzing} & 2019    & Data-Processing   & Multi-PIE, LFW, IJB-A      & Synthetic data pretraining followed by real data fine-tuning\\
Yucer \etal \cite{yucer2020exploring}            & 2020    & Data-Processing   & RFW & Adversarial image-to-image augmentation\\
Kotwal \& Marcel \cite{kotwal2024demographic}    & 2024    & Data-Processing   & RFW      & Demographic-dependent transformation of input image\\ 
\midrule
Amini \etal \cite{amini2019uncovering}           & 2019    & In-Processing     & PPB       & Sampling data probabilities for face detection\\
Alasadi \etal \cite{alasadi2019toward}           & 2019    & In-Processing     & CelebA, UMD Faces  & Introduces an adversarial learning framework using demographic classifier\\
Liang \etal\cite{liang2019additive}              & 2019    & In-Processing     & CelebA      & Two-stage adversarial bias mitigation through disentangled representations and additive adversarial learning\\
Wang \etal\cite{wang2019deep}                    & 2019    & In-Processing     & RFW     & Large-margin feature augmentation to balance class distributions\\
Yin \etal\cite{yin2019feature}                   & 2019    & In-Processing     & LFW, IJB-A, MS-Celeb-1M     & Feature transfer to enhance under-represented groups\\
Wang \& Deng \cite{wang2020mitigating}           & 2020    & In-Processing     & RFW        & Reinforcement learning-based race balance network\\
Gong \etal\cite{gong2020jointly}                 & 2020    & In-Processing     &  RFW       & Adversarial debiasing with identity and demographic classifiers\\
Georgopoulos \etal \cite{georgopoulos2020investigating} & 2020  & In-Processing  & KANFace            & Method to disentangle task-relevant features from bias terms in embeddings\\
Li \etal \cite{li2021learning}                   & 2021    & In-Processing     & RFW, BFW       & Progressive cross-transformer to remove race-induced identity-unrelated components\\
Gong \etal\cite{gong2021mitigating}              & 2021    & In-Processing     & RFW       & Group-adaptive training with adaptive convolution kernels and attention mechanisms\\
Xu \etal~\cite{xu2021consistent}                 & 2021    & In-Processing     & RFW, BFW          &   Additional penalty for disparities in FPR across instances\\
Wang \etal \cite{wang2021meta}                   & 2021    & In-Processing     & BUPT-Balancedface, Globalface & Meta Balanced Network (MBN): a bilevel optimization framework that learns adaptive margin values per skin-tone group\\
Park \etal \cite{park2022fair}                   & 2022    & In-Processing     & CelebA, UTK-Face    & Contrastive setups to enhance intra-class similarity and diminish similarity between negative samples\\
Serna \etal \cite{serna2022sensitive}            & 2022    & In-Processing     & DiveFace, RFW, BUPT-Balancedface & A triplet-based discrimination-aware loss, applicable to pre-trained FR models\\
Ma \etal \cite{Ma_2023_ICCV}                     & 2023    & In-Processing     & MFR, CelebA, RFW     &  Unsupervised data partitioning and Invariant Feature Regularization\\
Zhang \etal \cite{zhang2022fairness}             & 2023    & In-Processing     & CelebA, UTK-Face     & Generating contrastive sample pairs with visual similarity and unsupervised feature reweighting\\
Huang \etal \cite{huang2023gradient}             & 2023    & In-Processing     & RFW        & Enforces consistent gradient attention across demographic groups and follows attention-guided erasure to enhance feature learning\\
Wang \etal \cite{wang2023mixfairface}            & 2023    & In-Processing     & RFW    & Minimizes performance gaps across identities using MixFair Adapter- without relying on sensitive attribute labels\\
Dooley \etal~\cite{dooley2023rethinking}         & 2023    & In-processing     & VGGFace2, CelebA    & Uses Neural Architecture Search and Hyperparameter Optimization to discover fair FR architectures\\
Kotwal \& Marcel \cite{kotwal2024WACV}           & 2024    & In-Processing     & VGGFace2, MORPH, RFW & Regularization constraints based on score calibrations for demographic groups\\
Li \etal~\cite{li2025instance}                   & 2025    & In-Processing     & RFW, BFW, NFW       &   Customized instance margins to preserve consistency in FPR and TPR during the label classification in training\\
\bottomrule
\multicolumn{5}{>{\footnotesize\itshape}r}{Continued on the next page.}
\end{tabularx}
}
\end{table*}%
\begin{table*}[t]
\ContinuedFloat
\centering
\renewcommand{\arraystretch}{1.2}
\caption{\footnotesize{\it (Continued)}}
{\small
\begin{tabularx}{\linewidth}{l l l p{2cm} p{8cm}}
\toprule
\textbf{Reference}    & \textbf{Year} & \textbf{Type of Method} & \textbf{Test Dataset} & \textbf{Summary}\\ \midrule
Michalski \etal \cite{michalski2018impact}       & 2018    & Post-Processing   & Private      & Dynamic thresholds based on age differences\\ 
Srinivas \etal \cite{srinivas2019face}           & 2019    & Post-Processing   & ITWCC-D1       & Score fusion and ensemble strategies to address age-related bias\\
Robinson \etal \cite{robinson2020face}           & 2020    & Post-Processing   & BFW     & Demographic-specific thresholds\\
Terh\"{o}rst \etal \cite{terhorst2020comparison} & 2020    & Post-Processing   & Color-Feret, LFW  & Fairness-driven NN classifier\\
Morales \etal \cite{morales2020sensitivenets}    & 2020    & Post-Processing   & DiveFace, CelebA &  Disentangling sensitive attributes from features for privacy preservation\\
Terh\"{o}rst \etal\cite{terhorst2020post}        & 2020    & Post-Processing   & Color-Feret, MORPH   & Fair score normalization to mitigate demographic bias\\
Dhar \etal \cite{dhar2021pass}                   & 2021    & Post-Processing   & IJB-C    & Privacy-preserving disentanglement of features via adversarial training\\
Salvador \etal  \cite{salvador2021faircal}       & 2021    & Post-Processing   & RFW, BFW &  Conditional calibration of clustered embeddings\\
Conti \etal \cite{conti2022mitigating}           & 2022    & Post-Processing   & LFW  & Employing shallow MLP on the FR embeddings\\
Liu \etal \cite{liu2022oneface}                  & 2022    & Post-Processing   & RFW  & Threshold Consistency Penalty (TCP) to enforce consistent decision boundaries across multiple datasets\\
Robinson \etal \cite{robinson2023balancing}      & 2023    & Post-Processing   & BFW  & Privacy preserving domain adaptation framework reproject embeddings\\
Linghu \etal \cite{linghu2024ijcb}               & 2024    & Post-Processing   & VGGFace2, RFW    & Integrating demographic information in Z/T score normalization\\ 
Conti \& Cl{\'e}men{\c{c}}on \cite{conti2024mitigating} & 2024 & Post-processing & RFW     & Centroid-based loss aligns subgroup performance to a reference group via a lightweight fairness module\\
\bottomrule
\end{tabularx}
}
\end{table*}

Another approach for domain-specific bias mitigation using disentangled
representation learning was proposed by Liang \etal \cite{liang2019additive}.
They introduced a two-stage method combining modules for disentangled
representation learning with additive adversarial learning (AAL). While this
work does not directly address accuracy differential, it provides useful
insights into how domain-specific biases can be mitigated by learning
disentangled representations. The effectiveness of this method in reducing bias
across various domains suggests its potential applicability in the context of
demographic fairness in FR.
In \cite{alasadi2019toward}, the authors introduced an adversarial deep
learning framework for face matching aimed at reducing demographic bias. The
model incorporated a demographic classifier adversary during training to
encourage the learned feature representations to be invariant to the demographic
attribute (gender, in their work). Their results showed that the adversarial
setup reduced disparities in matching performance across demographic groups
while maintaining accuracy comparable to a standard baseline.

In \cite{li2021learning}, a Progressive Cross Transformer (PCT) was proposed to
mitigate racial bias by decoupling face representations into identity-related
and race-induced components. Using dual cross-transformers, the PCT refines
identity features and suppresses racial noise, demonstrating lower racial
differentials without compromising recognition accuracy.
In \cite{georgopoulos2020investigating}, the authors proposed a feature
decomposition-based framework that separates task-relevant features from
bias-related components in deep embeddings. In addition to FR, they applied this
framework to age estimation and gender classification, demonstrating that
certain demographic attributes can be disentangled without significantly
degrading task performance.

A bias mitigation strategy leveraging Neural Architecture Search (NAS)
and Hyperparameter Optimization (HPO) for fair FR models was proposed
in~\cite{dooley2023rethinking}. By constructing a tailored search space, their
authors synthesized architectures that generalize across datasets and protected
attributes, such as gender. The models resulted from this work demonstrated
reduced linear separability of demographic attributes.

The concept of score normalization was incorporated as a regularization term
into the training objective  enabling simultaneous optimization of recognition
accuracy and demographic fairness~\cite{kotwal2024WACV}. This was facilitated
by constraining the output scores of mated and non-mated pairs to adhere to a
pre-defined distribution, and followed by minimizing differences in score
distributions across demographic groups. During inference, the overall
pipeline did not require modifications as the FR CNN architecture remained
unaltered while only the weights were fine-tuned to the new objective.

A couple of works \cite{xu2021consistent, li2025instance} have proposed
customized instance margins aimed at preserving consistent error rates-- either
false positives or false positives combined with true positives. These methods
modify the Softmax-based loss by additional penalty terms derived from
disparities in the concerned error rates. Wang \etal \cite{wang2021meta}
proposed Meta Balanced Network (MBN), a meta-learning framework that dynamically
adjusted margin values for different skin tone groups using gradients from a
small, unbiased validation set. By framing margin optimization as an outer loop
and model training as an inner loop, MBN learnt adaptive margins that improve
fairness across groups. The bilevel optimization enabled their model to reduce
bias while maintaining performance. Serna \etal \cite{serna2022sensitive}
proposed Sensitive Loss, a discrimination-aware extension of the triplet loss
function that incorporated a sensitive triplet mining strategy. Designed as an
add-on to pre-trained FR models, this method enabled fairness improvements
without retraining entire architectures.
Wang \etal \cite{wang2023mixfairface} propose MixFairFace, a framework designed
to reduce identity bias-- performance inconsistencies across individual
identities-- rather than relying on demographic group labels. The core
component, MixFair Adapter, uses a sample-mixing strategy to estimate and
minimize identity-level bias during training. This label-free approach enables
fairness enhancement without requiring sensitive attribute annotations like race
or gender.

Finally, we discuss a couple of works dealing with fairness in facial attribute
recognition. Both of these are based on contrastive learning-- which can be
useful mechanism to address the demographic fairness in FR as well. Park \etal
addressed fairness issues in attribute classification using a contrastive
learning framework~\cite{park2022fair}. They constructed a Fair Supervised
Contrastive Loss (FSCL) which reduces disparities by normalizing intra-group
compactness and inter-group separability, penalizing sensitive attribute
information in representations. Although predominantly an in-processing method,
the authors also incorporate a loss function to address the imbalance in
training data where majority groups are constrained to have a better intra-
group compactness and inter-class separability compared to the
under-represented ones.
Similarly, Zhang \etal proposed Fairness-aware Contrastive Learning (FairCL) for
unsupervised representation learning and demonstrated the use-case of facial
attribute recognition \cite{zhang2022fairness}. In addition to fair contrastive
learning of feature representations, they also attempted to address the dataset
bias by specifically generating contrastive sample pairs that share the same
visual information apart from sensitive attributes. They also suggested
unsupervised feature reweighting to strike balance between the utility and
fairness of learned representations.

\subsection{Post-Processing Methods}

These methods are applied after the feature extraction or matching
stages to adjust decision scores and mitigate bias. Post-processing approaches
generally fall into two categories: those that modify the features
(\textit{embeddings}) and those that operate directly on the similarity scores
or their distributions. While these methods are less commonly used compared to
pre- and in-processing techniques, they can still play a role in ensuring
equitable outcomes. Additionally, post-processing methods are generally
easy to integrate into existing FR pipelines. Their advantages include avoiding
the need to retrain large models from scratch, enabling fairness adaptation
using a smaller dataset, faster training times, and the ability to retain the
performance of (a strong) pre-trained backbone. However, since the feature
extractor remains fixed, these methods may offer limited capacity to correct
upstream bias or improve learned representations.

Morales \etal \cite{morales2020sensitivenets} introduced SensitiveNets, a
framework that enforces the removal of sensitive attribute information from
learned feature representations. By disentangling sensitive attributes from
task-relevant features, their method ensures that downstream agents cannot
exploit protected information, thereby enhancing both privacy and fairness. 
Salvador \etal proposed FairCal: a post-processing method aimed at
improving fairness in face verification without retraining or requiring
sensitive attribute labels \cite{salvador2021faircal}. They clustered image
embeddings (obtained from frozen FR network) in an unsupervised manner and
applied conditional calibration maps per cluster. They reported reduced false
matches while improving overall accuracy. In \cite{conti2022mitigating}, Conti
\etal employed a shallow MLP on top of the frozen FR model as a post-processing
strategy to mitigate gender bias in FR. They enhanced the representations of
underperforming subgroups using a Fair von Mises-Fisher loss with class-specific
variance control. The work in \cite{conti2024mitigating} introduced a
post-processing method that improves fairness in pretrained (frozen) FR models
using a centroid-based loss called Centroid Fairness. A lightweight Fairness
Module was trained to align subgroup performance with that of a reference group,
without modifying the original model.

Feature descriptors, or \textit{embeddings}, generated by FR networks inherently
encode sensitive identity-related attributes, such as gender and skin-tone;
thereby rendering them vulnerable to privacy risks. Although the preservation of
privacy lies outside the scope of this review, it is closely linked to concerns
of fairness, given its role in exacerbating performance disparities across
demographic groups. Research on disentangling sensitive attributes from
identity-related representations remains relatively sparse. In
\cite{dhar2020towards}, Dhar \etal proposed the adversarial gender de-biasing
algorithm, which aims to reduce gender-specific information from embeddings
extracted by pre-trained (frozen) FR networks. This method employs adversarial
training to simultaneously suppress gender information and maintain identity
recognition performance. Building on this approach, the Protected Attribute
Suppression System (PASS)~\cite{dhar2021pass} advances the concept by
introducing a descriptor-level adversarial framework to mitigate accuracy
differentials. PASS operates directly on pre-trained embeddings and employs a
novel discriminator-based technique to suppress attributes such as gender and
skin tone without necessitating end-to-end retraining. Additionally, Robinson
\etal \cite{robinson2023balancing} highlighted that applying a single decision
threshold across demographic subgroups results in disparate performance,
diverging from global evaluation metrics. To address this, they proposed a
domain adaptation framework that re-projects embeddings into a feature space
where demographic attributes are suppressed, while identity-relevant information
is retained.

Michalski \etal investigated the impact of age variation on FR, particularly for
children~\cite{michalski2018impact}. They showed that dynamic thresholding
improves performance. To address age-related accuracy differences, they adjusted
thresholds based on age differences as opposed to a fixed threshold. In another
work on age-related bias, Srinivas \etal experimented with score-level fusion
strategies to improve recognition accuracy for the children (age as demographic)
\cite{srinivas2019face}. They considered six fusion schemes that combined
different score-normalization techniques and fusion rules. For normalization,
they considered $\mathbb{Z}$-norm and $\min$-$\max$ strategies; while fusion was
conducted using min, max, or sum rules.
Robinson \etal \cite{robinson2020face} showed that applying a single threshold
across different demographic groups leads to significant variations in the FMR.
They addressed this issue by using per-subgroup thresholds to balance the FMRs
across ethnic and gender groups, improving both recognition fairness and
performance. 

A typical FR pipeline employs similarity functions to obtain a matching score.
Terh{\"o}rst \etal replaced the conventional similarity function by a
fairness-driven neural network classifier \cite{terhorst2020comparison}. By
adding a penalization term in the loss function, their method was able to
equalizes score distributions across ethnic groups, reducing intra-ethnic bias
while maintaining high recognition performance. 
In another work \cite{terhorst2020post}, Terh{\"o}rst \etal introduced an
unsupervised fair score normalization method based on individual fairness
principles, which treats similar individuals similarly. During training, they
partitioned the identities in finite number of groups using K-means clustering
on face embeddings. At inference, they computed the cluster-specific thresholds
for both samples contributing to the score, and these threshold were combined
with a global threshold to yield normalized scores. 
In a recent work, Linghu \etal extended traditional score normalization methods,
such as $\mathbb{Z}$ and $\mathbb{T}$ normalization, by incorporating
demographic information to enhance fairness in FR systems~\cite{linghu2024ijcb}.
Furthermore, they evaluated three cohort-based approaches based on imposter
scores, Platt scaling, and bi-modal cumulative distribution functions (CDF).
Their findings demonstrated that the proposed method improved fairness across
both race and gender demographic groups, particularly at low FMRs.
 
%
Liu \etal \cite{liu2022oneface} argued that the existing FR evaluation
protocols, which rely on dataset-specific thresholds, are practical, and they
introduced the One-Threshold-for-All (OTA) protocol that employs a single
calibration threshold across all datasets. To optimize performance under the OTA
setting, they also proposed the Threshold Consistency Penalty (TCP), which
enforces consistent decision boundaries across domains through implicit domain
division followed by calibration and estimation of domain thresholds.

The works on mitigation are summarized in Table~\ref{tab:mitigation}. For most
of the state-of-the-art works aimed towards mitigating accuracy differentials,
the FR models have been trained on massive datasets, fine-tuned primarily for
accuracy, and are potentially operating near their capacity limits. When a new
objective, such as fairness, is introduced, it competes with the existing
accuracy objective, often resulting in performance degradation. This effect is
commonly known as the \textit{performance-fairness trade-off}, where
improvements in fairness are often accompanied by reduced recognition accuracy.

Some recent methods have demonstrated the potential to improve fairness metrics
without sacrificing overall accuracy. These methods, however, often involve
architectural modifications, such as the addition of auxiliary modules.  It is
important to note that in end-to-end FR pipelines, the total number of model
parameters can vary significantly depending on the chosen bias mitigation
strategy. Some in-processing methods aim to improve fairness by regularizing or
fine-tuning the existing FR network, without adding any additional modules. In
contrast, other approaches may integrate substantial additional components,
potentially increasing the network depth and capacity.  While such designs may
yield improved recognition performance along with reduced bias, they differ
significantly in architecture and complexity from the baseline models. In
addition to structural differences, variations in training data used for both--
bias mitigation and FR backbone-- play important role. As a result, performance
comparisons may not be directly equivalent, and should be made with transparency
and careful consideration.
%

\section{Future Directions}
\label{sec:future}

In recent years, with advanced architectures and increase in the number of
layers and parameters, FR models have gained a substantial improvement in their
capacity to learn complex facial representations. These deeper architectures,
often comprising over a hundred layers with millions of parameters, have
significantly enhanced the ability of FR systems to generalize across
challenging scenarios, resulting in higher overall recognition performance (as
well as improved fairness). Additionally, the availability of larger, more
diverse datasets has contributed to better learning outcomes. These datasets,
which incorporate substantial variation in demographics and covariates such as
pose, illumination, and expression (PIE), have facilitated measurable progress
in improving both accuracy and fairness FR.

However, despite these advancements, several key challenges in this area
continue to exist. Most of the existing efforts primarily center on extremely
deep models, which demand extensive computational resources and memory footprint
for both training and deployment. This emphasis on high-capacity architectures
does not adequately address quality issues in data or labels nor does it cater
to the requirements of resource-constrained environments. Thus, while the
combination of deeper models and diverse data has been pivotal, future research
must explore avenues to address residual biases and expand fairness to a broader
range of applications.  

In this section, we examine some of the emerging challenges associated with
demographic fairness. These challenges highlight the need for ongoing research to
adapt bias mitigation strategies to align with the advancements in FR
applications.\\

\noindent\textbf{Lightweight Models:} The lightweight FR models, often used in
handheld devices and resource-constrained environments, encounter significant
challenges concerning demographic fairness. These systems, crucial for
privacy-sensitive applications, often inherit limitations in capacity and
architecture, leading to non-equitable performance across demographic groups.
Performance disparities in lightweight models has garnered limited research
attention, despite their widespread deployment in mobile and IoT devices with
varying sensor qualities. Techniques like knowledge distillation (KD) and
pruning, while essential for model compression, introduce or amplify bias. For
instance, Liu \etal highlighted that KD inherits biases from larger teacher
models \cite{liu2021rectifying}, while pruning has been shown to
disproportionately impacts underrepresented groups~\cite{paganini2020prune,
iofinova2023bias}.
Incorporating fairness-aware techniques is crucial for mitigating these issues.
Lin \etal \cite{lin2022fairgrape} introduced FairGRAPE, a pruning method that
evaluates network connections with demographic considerations, reducing
performance disparities. Caldeira \etal \cite{caldeira2024mst} proposed a
multi-specialized teacher framework where each teacher model is trained on a
specific ethnicity and collectively distills knowledge into a student model.
Achieving demographic fairness in lightweight models requires targeted
compression strategies and ethical evaluations of demographic-specific impacts.
By integrating fairness principles into compression techniques, lightweight FR
systems can achieve more equitable outcomes while maintaining efficiency and
accuracy.\\

\noindent\textbf{Quantization of Models and Data:} Quantization, a model
compression technique, retains the original architecture (as opposed to KD and
pruning-- which often modify the structure) while reducing parameter precision,
producing smaller and faster models. However, it can lead to higher disparities
across demographic groups by prioritizing global performance over the accurate
classification of under-represented groups. Such performance disparities
underscore the need for rigorous fairness evaluation across demographic
subgroups when deploying compressed models.
Quantization converts floating-point (FP) models into lower-precision formats
like 8-bit, balancing efficiency and accuracy~\cite{jacob2018quantization}. It
typically consists of two approaches: post-training quantization (PTQ) and
quantization-aware training (QAT). Some studies, such as Stoychev \etal
demonstrated that 8-bit PTQ maintained fairness and accuracy in gender bias for
face expression recognition \cite{jacob2018quantization}. However, reducing
precision to 6 bits significantly degraded fairness, indicating a trade-off
between compression and bias mitigation. Although similar investigations for FR
systems remain limited, these findings highlight the need to evaluate and ensure
demographic fairness in quantized models.

While the majority of research in this area focuses on quantized models, a
number of recent studies have investigated the implications of lossy image
compression on fairness in FR. The work in \cite{yucer2022does} explored how
lossy JPEG compression influences recognition accuracy across demographic
groups, more specifically on racial phenotypes. Their findings indicate that
compression disproportionately degrades performance for certain traits, such as
darker skin-tones and wider facial features. They also demonstrated that
disabling chroma sub-sampling can improve FMRs across several racial phenotypes.
Similarly, Qiu \etal \cite{qiu2024gone} examined the effects of five neural
image compression methods and found that certain phenotypic features--such as
skin-tone and hair-type--are more susceptible to degradation at lower bit rates.
Notably, at extremely low bit rates, the amplification of bias was particularly
pronounced for individuals belonging to African race.\\

\noindent\textbf{Low Resolution:} The research on demographic fairness in FR has
predominantly focused on high-resolution images, often overlooking the
challenges posed by low-resolution images typically captured by surveillance
cameras or from significant distances.  One of the primary impediment to
research this issue is the scarcity of low-resolution datasets that include
demographic attributes. Consequently, demographic disparities in low-resolution
FR remain under-explored, despite their importance in real-world applications.
A recent work from Atzori \etal attempted to address this gap by designing a
novel framework to investigate performance differences across demographic groups
in low-resolution FR \cite{atzori2023demographic}. They trained state-of-the-art
FR models on various combinations of high- and low-resolution images. Testing on
degraded images from five datasets revealed significant disparities across
gender and ethnic groups, underscoring the need for timely interventions in
low-resolution FR. It may be noted that their approach involved use of a
generative model to convert high-resolution face images into realistic
low-resolution counterparts.
In \cite{bhatta2024impact, pangelinan2024analyzing}, the impact of blur and
image resolution were investigated in the context of 1:$N$ identification,
revealing that increased blurring leads to elevated FPIRs. These studies also
noted that the impact of blur varies across demographic attributes, particularly
race and gender. The quality degradations of blur and resolution in these
studies were simulated using Gaussian blurring and bicubic interpolation,
respectively.
The importance of low-resolution FR is evident in programs like
BRIAR\footnote{\href{https://www.iarpa.gov/research-programs/briar}{BRIAR
Programme} by IARPA.}, which aim to enhance recognition technologies for
challenging scenarios, such as long-distance identification and low-quality
image acquisition. To tackle demographic fairness in low-resolution FR, there is
a need to develop both datasets and models tailored to these unique use-cases.\\

\noindent\textbf{Datasets for Training and Evaluation:} While demographically
balanced training datasets alone may not fully eliminate performance differences
in FR, they play a critical role in reducing such disparities. Achieving
fairness requires not only demographic diversity (\eg across race, gender, and
age) but also balance with respect to non-demographic or soft attributes such as
hairstyle, facial hair, makeup, and occlusion. Therefore, large-scale datasets
that are balanced across both demographic and non-demographic dimensions are
essential for developing FR models that are both fair and accurate. However,
acquiring such datasets is increasingly challenging due to cost and ethical and
privacy concerns surrounding biometric data collection. The commercial sector,
in particular, faces difficulties as most available datasets are collected
locally from consenting individuals---which are often in limited size and
demographic representation. These constraints necessitate innovative approaches
to address demographic fairness using smaller or synthetic datasets.
When it comes to evaluation and benchmarking, many existing datasets are
significantly imbalanced-- typically over-representing a single demographic
group, such as Caucasians. Alongside developing more inclusive training
datasets, there is a pressing need to construct evaluation datasets that more
accurately reflect demographic diversity.

Access to datasets annotated with multiple attributes--both demographic and
non-demographic--will prove to be a valuable resource for advancing fairness in
biometric recognition. Such datasets enable deeper exploration of the underlying
causes of performance differences across demographic groups by facilitating the
analysis of correlations and potential causal relationships among factors that
influence recognition performance. Moreover, they provide a foundation for
designing effective mitigation strategies. When used for assessment,
multi-attribute datasets can help controlling or isolating individual factors,
enabling systematic benchmarking of their specific impact on model performance.

The use of synthetic data in FR has recently gained traction as a potential
solution to privacy and data-sharing concerns. Competitions like the FRSyn
series have encouraged advancements in synthetic data
usage~\cite{melzi2024frcsyn1, deandres2024frcsyn, melzi2024frcsyn}. Despite
these efforts, FR models trained exclusively on synthetic datasets continue to
underperform compared to those trained on real datasets of similar
size~\cite{melzi2024frcsyn, george2024digi2real}. This gap is evident in both
recognition accuracy (measured by metrics like FMR and FNMR) and demographic
fairness (assessed by standard deviation of performance across groups). The
analysis of synthetic datasets by Huber \etal revealed that demographic bias
might worsen compared to the (real) training dataset~\cite{huber2024bias}.
Enhancing the quality and utility of synthetic datasets, beyond the aspect of
fairness, remains an open problem, requiring further exploration.\\

\noindent\textbf{Causes of Performance Disparities:} Recent research indicates
that accuracy disparities across demographic groups in FR systems are not solely
attributable to inherent physical or biological characteristics. Instead, these
disparities are frequently influenced by non-demographic factors-- such as
hairstyle, cosmetic use, or occlusion patterns-- that are shaped by cultural
norms, social behaviors, or individual preferences. Despite their relevance,
such factors have historically received limited attention, with only a few
recent studies beginning to reveal their significant impact on algorithmic
performance.

A comprehensive understanding of the origins of demographic accuracy
differentials requires disentangling the complex interplay between demographic
attributes and confounding behavioral or environmental variables. This
necessitates the development of training and evaluation datasets that are not
only demographically diverse but also annotated for a broader set of
non-demographic characteristics potentially correlated with recognition
performance.

To systematically analyze these influences, it is essential to design datasets
that enable controlled isolation-- analogous to evaluating a partial
derivative-- of individual or combined factors. This would allow researchers to
measure the effect of specific attributes or conditions on the performance of a
FR model or pipeline. 

Advancing our understanding of these causal relationships is critical not only
for identifying the root causes of observed biases but also for facilitating
systematic and reproducible evaluation. Ultimately, understanding these
underlying causes is a prerequisite for developing effective mitigation
strategies and building FR systems that are both fair and accurate.\\

\noindent\textbf{Use-Cases of Remote Checking:} Last few years have witnessed
tremendous surge in online activities: financial transactions, banking,
user-onboarding, etc. These activities have driven widespread adoption of remote
identity verification (RIdV) technologies. These systems authenticate
individuals by comparing real-time images or selfies, captured via smart
devices, against official identity documents, such as work permits or driver's
licenses. Such solutions are integral to online Know Your Customer (KYC)
processes, which are now standard for banks and financial institutions. While
RIdV systems enhance convenience and scalability, it is essential to ensure
their fairness across demographic groups as they become more prevalent.

Recognizing the increasing reliance on remote verification, the MdTF and DHS
S\&T introduced the Remote Identity Validation Technology Demonstration (RIVTD)
initiative\footnote{\href{https://mdtf.org/rivtd}{Remote Identity Validation
Technology Demonstration (RIVTD)}}. In addition to security, accuracy, and
liveness detection requirements, this program also places particular emphasis on
ensuring demographic fairness in such technologies.
A recent study by Fatima \etal\cite{fatima2024large} investigated demographic
fairness in RIdV technologies using statistical methods to analyze performance
disparities. Their analysis of five commercial RIdV systems revealed that only
two achieved equitable outcomes across demographic groups. Notably, higher FNMRs
were observed among African cohorts and individuals with darker skin-tones. Such
findings highlight the necessity of evaluating RIdV technologies across
demographic groups to ensure equitable and fair performance.\\

\noindent\textbf{Complex Bias Factors (Intersectionality):} The majority of
research on mitigation of accuracy differentials in FR, as discussed in
Sec~\ref{sec:mitigation} has focused on single demographic attributes, such as
race, age, or gender. However, several studies have identified that combination
or intersection of various demographic factors causes (or amplifies) bias in FR
models (cf. Sec~\ref{sec:causes}), while few works, such as \cite{Liu_2022_CVPR,
pangelinan2025lights} have addressed the issue of disparities caused due to
multiple or combined attributes. Existing bias mitigation techniques typically
target one demographic attribute at a time, achieving measurable improvements in
fairness for that specific attribute. However, in most cases, it remains
unclear whether such processing inadvertently introduces imbalances in other
demographic attributes. For instance, enhancing fairness for gender-related bias
may increase disparities linked to ethnicity and vice-versa. This highlights the
need for systematic evaluations of the intersectionality of demographic factors,
such as race and gender combined. Consequently, developing mitigation methods
capable of addressing multiple demographic attributes simultaneously remains an
open challenge.\\

\noindent\textbf{Noisy Labels:} The assignment of demographic attributes such as
race, ethnicity, and skin-tone in FR datasets typically involves discrete
labeling into finite categories. Some attributes, such as race, are often
self-reported. In many cases, race and ethnicity annotations may be derived from
automatic classifiers or manual efforts. Automatic classifiers, predominantly
based on deep learning, are likely to be susceptible to bias too; while manual
annotations are prone to human judgment errors. Similarly, skin-tone is
frequently categorized using scales like Fitzpatrick's, which discretizes it
into specific values, ignoring its continuous spectrum. This reliance on
discrete labels introduces noise into the training data, as samples near
category boundaries are often inaccurately labeled. Recent work
\cite{kotwal2024demographic}, addressed this issue by using probabilistic
weights (soft labels) for demographic information instead of utilizing rigid
(categorical) labels. However, most existing methods overlook the issue of
errors in training data.

Noisy labels in training datasets pose a significant challenge, especially
considering massive scale of FR datasets, often in few hundreds of thousands of
images. Manually verifying or curating such datasets is labor-intensive,
impractical, and still prone to errors. Furthermore, removing samples with
ambiguous labels can lead to reducing dataset diversity and robustness. Thus,
developing robust mitigation strategies capable of handling noisy labels without
compromising the effectiveness of training processes or dataset diversity is
essential for improving fairness and accuracy in FR systems.
%
%

\section{Conclusion}
\label{sec:conc}

In this work, we have systematically explored the issue of performance
differences in FR across demographic groups, or \textit{bias} through different
yet interrelated sections: causes, datasets, assessment metrics, and mitigation
strategies. We discussed key contributing factors such as training data
imbalance, skin-tone variations, and image quality, as well as the growing
recognition of non-demographic attributes. Our analysis of datasets emphasized
the importance of demographic diversity and the need for multi-attribute
annotations to enable deeper understanding of performance disparities. We also
reviewed a range of fairness evaluation metrics and mitigation techniques across
the FR pipeline.

As highlighted by several recent studies, non-demographic covariates (or
soft-biometric factors) can have a significant influence on FR performance,
including accuracy and error rates. Given that both training and evaluation
datasets, including real-world deployments, are rarely controlled for
demographic attributes in isolation, it becomes critical to reconsider how we
interpret variations in FR outcomes across demographic groups. Many of these
non-demographic factors are closely entangled with demographic attributes (\eg
hairstyles and makeup often vary by culture or gender). Consequently,
performance disparities observed across groups may not be attributable to the
demographic variable alone. Without precise isolation and identification of
contributing factors, attributing lower FR performance to demographic bias may
be misleading. Therefore, caution must be exercised before concluding that an FR
system exhibits bias toward a particular demographic group.

Another important challenge to be noted here is the trade-off between fairness
and accuracy. Since most FR models are already optimized for performance, adding
fairness objectives often introduces conflicting constraints. Although some
architectures sometimes improve both, these gains frequently rely on increased
model capacity, rather than an inherent resolution of the fairness-accuracy
trade-off. Recent studies have shown that gender-related performance differences
may be explained by soft-biometric attributes rather than inherent biological
differences, suggesting that some disparities are socio-cultural in nature.
However, such understanding is still limited for attributes like race or
skin-tone, largely due to the absence of datasets annotated with both
demographic and non-demographic variables. Creating such datasets is essential
but resource-intensive; synthetic data may help address this gap, though its
effectiveness remains sub-par of real-world data. Improved metrics are also
needed to quantify the impact of specific attributes on performance in a
consistent and interpretable way. Additionally, most fairness evaluations are
conducted on large-scale models, yet future deployments--especially on mobile
and edge devices--will require lightweight models that may be more prone to
differential treatment. Ensuring fairness under such constraints is critical for
real-world adoption.

Together, these challenges reflect the evolving nature of FR technologies and
underscore the need for innovative strategies to ensure equitable and reliable
outcomes in real-world applications.
%


\section*{Acknowledgment}
The work has been supported by the Hasler foundation (through the SAFER project)
and the Swiss Center for Biometrics Research and Testing.

\balance
\bibliographystyle{IEEEtran}
\bibliography{main}

\begin{thebibliography}{100}
\providecommand{\url}[1]{#1}
\csname url@samestyle\endcsname
\providecommand{\newblock}{\relax}
\providecommand{\bibinfo}[2]{#2}
\providecommand{\BIBentrySTDinterwordspacing}{\spaceskip=0pt\relax}
\providecommand{\BIBentryALTinterwordstretchfactor}{4}
\providecommand{\BIBentryALTinterwordspacing}{\spaceskip=\fontdimen2\font plus
\BIBentryALTinterwordstretchfactor\fontdimen3\font minus
  \fontdimen4\font\relax}
\providecommand{\BIBforeignlanguage}[2]{{%
\expandafter\ifx\csname l@#1\endcsname\relax
\typeout{** WARNING: IEEEtran.bst: No hyphenation pattern has been}%
\typeout{** loaded for the language `#1'. Using the pattern for}%
\typeout{** the default language instead.}%
\else
\language=\csname l@#1\endcsname
\fi
#2}}
\providecommand{\BIBdecl}{\relax}
\BIBdecl

\bibitem{rathgeb2022demographic}
C.~Rathgeb, P.~Drozdowski, D.~Frings, N.~Damer, and C.~Busch, ``Demographic
  fairness in biometric systems: What do the experts say?'' \emph{IEEE
  Technology and Society Magazine}, vol.~41, no.~4, pp. 71--82, 2022.

\bibitem{sixta2020fairface}
T.~Sixta, J.~C. Jacques~Junior, P.~Buch-Cardona, E.~Vazquez, and S.~Escalera,
  ``Fairface challenge at eccv 2020: Analyzing bias in face recognition,'' in
  \emph{Computer Vision--ECCV 2020 Workshops: Glasgow, UK, August 23--28, 2020,
  Proceedings, Part VI 16}.\hskip 1em plus 0.5em minus 0.4em\relax Springer,
  2020, pp. 463--481.

\bibitem{drozdowski2020demographic}
P.~Drozdowski, C.~Rathgeb, A.~Dantcheva, N.~Damer, and C.~Busch, ``Demographic
  bias in biometrics: A survey on an emerging challenge,'' \emph{IEEE
  Transactions on Technology and Society}, vol.~1, no.~2, pp. 89--103, 2020.

\bibitem{jain2021biometrics}
A.~Jain, D.~Deb, and J.~Engelsma, ``Biometrics: Trust, but verify,'' \emph{IEEE
  Transactions on Biometrics, Behavior, and Identity Science}, vol.~4, no.~3,
  pp. 303--323, 2021.

\bibitem{hill2020wrongfully}
\BIBentryALTinterwordspacing
K.~Hill, ``Wrongfully accused by an algorithm,'' \emph{The New York Times},
  2020. [Online]. Available:
  \url{https://www.nytimes.com/2020/06/24/technology/facial-recognition-arrest.html}
\BIBentrySTDinterwordspacing

\bibitem{aclu2018rekognition}
\BIBentryALTinterwordspacing
J.~Snow, ``Amazon's face recognition falsely matched 28 members of congress
  with mugshots,'' \emph{ACLU Blog}, 2018. [Online]. Available:
  \url{https://www.aclu.org/blog/privacy-technology/surveillance-technologies/amazons-face-recognition-falsely-matched-28}
\BIBentrySTDinterwordspacing

\bibitem{marshall2023why}
\BIBentryALTinterwordspacing
L.~Marshall, ``Why new facial-recognition airport screenings are raising
  concerns,'' 2023. [Online]. Available:
  \url{https://www.colorado.edu/today/2023/07/11/why-new-facial-recognition-airport-screenings-are-raising-concerns}
\BIBentrySTDinterwordspacing

\bibitem{falk2024struggle}
\BIBentryALTinterwordspacing
E.~Falk, ``The struggle to control facial recognition at airports,'' 2024.
  [Online]. Available:
  \url{https://journals.library.columbia.edu/index.php/stlr/blog/view/607}
\BIBentrySTDinterwordspacing

\bibitem{mehrabi2021survey}
N.~Mehrabi, F.~Morstatter, N.~Saxena, K.~Lerman, and A.~Galstyan, ``A survey on
  bias and fairness in machine learning,'' \emph{ACM computing surveys (CSUR)},
  vol.~54, no.~6, pp. 1--35, 2021.

\bibitem{howard2019effect}
J.~J. Howard, Y.~B. Sirotin, and A.~R. Vemury, ``The effect of broad and
  specific demographic homogeneity on the imposter distributions and false
  match rates in face recognition algorithm performance,'' in \emph{2019 IEEE
  10th International Conference on Biometrics Theory, Applications and Systems
  (BTAS)}.\hskip 1em plus 0.5em minus 0.4em\relax IEEE, 2019, pp. 1--8.

\bibitem{limante2024}
A.~Limant{\.e}, ``Bias in facial recognition technologies used by law
  enforcement: Understanding the causes and searching for a way out,''
  \emph{Nordic Journal of Human Rights}, vol.~42, no.~2, pp. 115--134, 2024.

\bibitem{jones2020law}
C.~Jones, ``Law enforcement use of facial recognition: bias, disparate impacts
  on people of color, and the need for federal legislation,'' \emph{NCJL \&
  Tech.}, vol.~22, p. 777, 2020.

\bibitem{leslie2020understanding}
D.~Leslie, ``Understanding bias in facial recognition technologies,''
  \emph{arXiv preprint arXiv:2010.07023}, 2020.

\bibitem{burgess2022watching}
B.~Burgess, A.~Ginsberg, E.~W. Felten, and S.~Cohney, ``Watching the watchers:
  bias and vulnerability in remote proctoring software,'' in \emph{31st USENIX
  security symposium (USENIX security 22)}, 2022, pp. 571--588.

\bibitem{raji2020saving}
I.~D. Raji, T.~Gebru, M.~Mitchell, J.~Buolamwini, J.~Lee, and E.~Denton,
  ``Saving face: Investigating the ethical concerns of facial recognition
  auditing,'' in \emph{Proceedings of the AAAI/ACM Conference on AI, Ethics,
  and Society}, 2020, pp. 145--151.

\bibitem{pena2020bias}
A.~Pena, I.~Serna, A.~Morales, and J.~Fierrez, ``Bias in multimodal ai: Testbed
  for fair automatic recruitment,'' in \emph{Proceedings of the IEEE/CVF
  Conference on Computer Vision and Pattern Recognition Workshops}, 2020, pp.
  28--29.

\bibitem{ross2019some}
A.~Ross, S.~Banerjee, C.~Chen, A.~Chowdhury, V.~Mirjalili, R.~Sharma,
  T.~Swearingen, and S.~Yadav, ``Some research problems in biometrics: The
  future beckons,'' in \emph{2019 International Conference on Biometrics
  (ICB)}.\hskip 1em plus 0.5em minus 0.4em\relax IEEE, 2019, pp. 1--8.

\bibitem{busch2024challenges}
C.~Busch, ``Challenges for automated face recognition systems,'' \emph{Nature
  Reviews Electrical Engineering}, pp. 1--10, 2024.

\bibitem{frvt3}
P.~Grother, M.~Ngan, and K.~Hanaoka, ``Face recognition vendor test part 3:
  Demographic effects,'' 12 2019.

\bibitem{grother2021demographic}
P.~Grother, ``Demographic differentials in face recognition algorithms,''
  \emph{EAB Virtual Event Series-Demographic Fairness in Biometric Systems},
  2021.

\bibitem{9833480}
W.~Deng, T.~Hassner, X.~Liu, and M.~Pantic, ``Tbiom special issue on
  trustworthy biometrics-editorial,'' \emph{IEEE Transactions on Biometrics,
  Behavior, and Identity Science}, vol.~4, no.~3, pp. 301--302, 2022.

\bibitem{cheong2021hitchhiker}
J.~Cheong, S.~Kalkan, and H.~Gunes, ``The hitchhiker's guide to bias and
  fairness in facial affective signal processing: Overview and techniques,''
  \emph{IEEE Signal Processing Magazine}, vol.~38, no.~6, pp. 39--49, 2021.

\bibitem{british2023iso}
{ISO/IEC JTC1 SC37 Biometrics}, ``{ISO/IEC DIS 19795-10. Information Technology
  -- Biometric Performance Testing and Reporting -- Part 10: Quantifying
  Biometric System Performance Variation Across Demographic Group},'' 2023.

\bibitem{pessach2022review}
D.~Pessach and E.~Shmueli, ``A review on fairness in machine learning,''
  \emph{ACM Computing Surveys (CSUR)}, vol.~55, no.~3, pp. 1--44, 2022.

\bibitem{yucer2023racial}
S.~Yucer, F.~Tektas, N.~Al~Moubayed, and T.~Breckon, ``Racial bias within face
  recognition: A survey,'' \emph{ACM Computing Surveys}, 2023.

\bibitem{krishnapriya2020issues}
K.~Krishnapriya, V.~Albiero, K.~Vangara, M.~C. King, and K.~W. Bowyer, ``Issues
  related to face recognition accuracy varying based on race and skin tone,''
  \emph{IEEE Transactions on Technology and Society}, vol.~1, no.~1, pp. 8--20,
  2020.

\bibitem{klare2012face}
B.~F. Klare, M.~J. Burge, J.~C. Klontz, R.~W.~V. Bruegge, and A.~K. Jain,
  ``Face recognition performance: Role of demographic information,'' \emph{IEEE
  Transactions on Information Forensics and Security}, vol.~7, no.~6, pp.
  1789--1801, 2012.

\bibitem{cavazos2020accuracy}
J.~G. Cavazos, P.~J. Phillips, C.~D. Castillo, and A.~J. O'Toole, ``Accuracy
  comparison across face recognition algorithms: Where are we on measuring race
  bias?'' \emph{IEEE Transactions on Biometrics, Behavior, and Identity
  Science}, vol.~3, no.~1, pp. 101--111, 2020.

\bibitem{gwilliam2021rethinking}
M.~Gwilliam, S.~Hegde, L.~Tinubu, and A.~Hanson, ``Rethinking common
  assumptions to mitigate racial bias in face recognition datasets,'' in
  \emph{Proceedings of the IEEE/CVF International Conference on Computer
  Vision}, 2021, pp. 4123--4132.

\bibitem{albiero2020does}
V.~Albiero, K.~Bowyer, K.~Vangara, and M.~King, ``Does face recognition
  accuracy get better with age? deep face matchers say no,'' in
  \emph{Proceedings of the IEEE/CVF Winter Conference on Applications of
  Computer Vision}, 2020, pp. 261--269.

\bibitem{albiero2020does_ijcb}
V.~Albiero, K.~Zhang, and K.~W. Bowyer, ``How does gender balance in training
  data affect face recognition accuracy?'' in \emph{2020 IEEE International
  Joint Conference on Biometrics (IJCB)}.\hskip 1em plus 0.5em minus
  0.4em\relax IEEE, 2020, pp. 1--10.

\bibitem{wu2023should}
H.~Wu and K.~W. Bowyer, ``What should be balanced in a ``balanced'' face
  recognition dataset?'' \emph{arXiv preprint arXiv:2304.09818}, 2023.

\bibitem{donida2022soft}
R.~Donida~Labati, A.~Ross, and A.~Dantcheva, ``Soft biometrics,'' in
  \emph{Encyclopedia of Cryptography, Security and Privacy}.\hskip 1em plus
  0.5em minus 0.4em\relax Springer, 2022, pp. 1--4.

\bibitem{wang2019racial}
M.~Wang, W.~Deng, J.~Hu, X.~Tao, and Y.~Huang, ``Racial faces in the wild:
  Reducing racial bias by information maximization adaptation network,'' in
  \emph{Proceedings of the IEEE/CVF International Conference on Computer
  Vision}, 2019, pp. 692--702.

\bibitem{kolla2023impact}
M.~Kolla and A.~Savadamuthu, ``The impact of racial distribution in training
  data on face recognition bias: A closer look,'' in \emph{Proceedings of the
  IEEE/CVF Winter Conference on Applications of Computer Vision}, 2023, pp.
  313--322.

\bibitem{muthukumar2018understanding}
V.~Muthukumar, T.~Pedapati, N.~Ratha, P.~Sattigeri, C.-W. Wu, B.~Kingsbury,
  A.~Kumar, S.~Thomas, A.~Mojsilovic, and K.~R. Varshney, ``Understanding
  unequal gender classification accuracy from face images,'' \emph{arXiv
  preprint arXiv:1812.00099}, 2018.

\bibitem{cook2019demographic}
C.~M. Cook, J.~J. Howard, Y.~B. Sirotin, J.~L. Tipton, and A.~R. Vemury,
  ``Demographic effects in facial recognition and their dependence on image
  acquisition: An evaluation of eleven commercial systems,'' \emph{IEEE
  Transactions on Biometrics, Behavior, and Identity Science}, vol.~1, no.~1,
  pp. 32--41, 2019.

\bibitem{fitzpatrick1988validity}
T.~B. Fitzpatrick, ``The validity and practicality of sun-reactive skin types i
  through vi,'' \emph{Archives of dermatology}, vol. 124, no.~6, pp. 869--871,
  1988.

\bibitem{monk2023monk}
E.~Monk, ``The monk skin tone scale,'' \emph{OSF}, 2023.

\bibitem{buolamwini18a}
J.~Buolamwini and T.~Gebru, ``Gender shades: Intersectional accuracy
  disparities in commercial gender classification,'' in \emph{Proceedings of
  the Conference on Fairness, Accountability and Transparency}, ser.
  Proceedings of Machine Learning Research, S.~Friedler and C.~Wilson, Eds.,
  vol.~81, Feb 2018, pp. 77--91.

\bibitem{lu2019experimental}
B.~Lu, J.-C. Chen, C.~D. Castillo, and R.~Chellappa, ``An experimental
  evaluation of covariates effects on unconstrained face verification,''
  \emph{IEEE Transactions on Biometrics, Behavior, and Identity Science},
  vol.~1, no.~1, pp. 42--55, 2019.

\bibitem{phillips2011other}
P.~J. Phillips, F.~Jiang, A.~Narvekar, J.~Ayyad, and A.~J. O'Toole, ``An
  other-race effect for face recognition algorithms,'' \emph{ACM Transactions
  on Applied Perception (TAP)}, vol.~8, no.~2, pp. 1--11, 2011.

\bibitem{ricanek2015review}
K.~Ricanek, S.~Bhardwaj, and M.~Sodomsky, ``A review of face recognition
  against longitudinal child faces,'' \emph{BIOSIG 2015}, pp. 15--26, 2015.

\bibitem{nagpal2019deep}
S.~Nagpal, M.~Singh, R.~Singh, and M.~Vatsa, ``Deep learning for face
  recognition: Pride or prejudiced?'' \emph{arXiv preprint arXiv:1904.01219},
  2019.

\bibitem{albiero2020analysis}
V.~Albiero, K.~Krishnapriya, K.~Vangara, K.~Zhang, M.~C. King, and K.~W.
  Bowyer, ``Analysis of gender inequality in face recognition accuracy,'' in
  \emph{Proceedings of the IEEE/CVF Winter Conference on Applications of
  Computer Vision workshops}, 2020, pp. 81--89.

\bibitem{dooley2023rethinking}
S.~Dooley, R.~Sukthanker, J.~Dickerson, C.~White, F.~Hutter, and M.~Goldblum,
  ``Rethinking bias mitigation: Fairer architectures make for fairer face
  recognition,'' \emph{Advances in Neural Information Processing Systems},
  vol.~36, pp. 74\,366--74\,393, 2023.

\bibitem{vangara2019characterizing}
K.~Krishnapriya, K.~Vangara, M.~C. King, V.~Albiero, K.~Bowyer \emph{et~al.},
  ``Characterizing the variability in face recognition accuracy relative to
  race,'' in \emph{Proceedings of the IEEE/CVF Conference on Computer Vision
  and Pattern Recognition Workshops}, 2019, pp. 0--0.

\bibitem{cook_tbiom}
C.~Cook, J.~Howard, Y.~Sirotin, J.~Tipton, and A.~Vemury, ``Demographic effects
  in facial recognition and their dependence on image acquisition: An
  evaluation of eleven commercial systems,'' \emph{IEEE Transactions on
  Biometrics, Behavior, and Identity Science}, vol.~1, no.~1, pp. 32--41, 2019.

\bibitem{cook2023demographic}
C.~M. Cook, J.~J. Howard, Y.~B. Sirotin, J.~L. Tipton, and A.~R. Vemury,
  ``Demographic effects across 158 facial recognition systems,'' Technical
  report, DHS, Tech. Rep., 2023.

\bibitem{wu2023face}
H.~Wu, V.~Albiero, K.~Krishnapriya, M.~King, and K.~Bowyer, ``Face recognition
  accuracy across demographics: Shining a light into the problem,'' in
  \emph{Proceedings of the IEEE/CVF Conference on Computer Vision and Pattern
  Recognition}, 2023, pp. 1041--1050.

\bibitem{bhatta2024impact}
A.~Bhatta, G.~Pangelinan, M.~C. King, and K.~W. Bowyer, ``Impact of blur and
  resolution on demographic disparities in 1-to-many facial identification,''
  in \emph{Proceedings of the IEEE/CVF Winter Conference on Applications of
  Computer Vision}, 2024, pp. 412--420.

\bibitem{pangelinan2025lights}
G.~Pangelinan, G.~Bezold, H.~Wu, M.~King, and K.~Bowyer, ``Lights, camera,
  matching: The role of image illumination in fair face recognition,''
  \emph{arXiv preprint arXiv:2501.08910}, 2025.

\bibitem{el2016face}
H.~El~Khiyari and H.~Wechsler, ``Face verification subject to varying (age,
  ethnicity, and gender) demographics using deep learning,'' \emph{Journal of
  Biometrics and Biostatistics}, vol.~7, no. 323, p.~11, 2016.

\bibitem{best2017longitudinal}
L.~Best-Rowden and A.~K. Jain, ``Longitudinal study of automatic face
  recognition,'' \emph{IEEE Transactions on Pattern Analysis and Machine
  Intelligence}, vol.~40, no.~1, pp. 148--162, 2017.

\bibitem{vera2019facegenderid}
R.~Vera-Rodriguez, M.~Blazquez, A.~Morales, E.~Gonzalez-Sosa, J.~C. Neves, and
  H.~Proen{\c{c}}a, ``Facegenderid: Exploiting gender information in dcnns face
  recognition systems,'' in \emph{Proceedings of the IEEE/CVF Conference on
  Computer Vision and Pattern Recognition Workshops}, 2019, pp. 0--0.

\bibitem{fr_age}
V.~Albiero, K.~Bowyer, K.~Vangara, and M.~C. King, ``Does face recognition
  accuracy get better with age? deep face matchers say no,'' in
  \emph{Proceedings of the IEEE Winter Conference on Applications of Computer
  Vision}, vol.~1, 2020, pp. 250--258.

\bibitem{atzori2022explaining}
A.~Atzori, G.~Fenu, and M.~Marras, ``Explaining bias in deep face recognition
  via image characteristics,'' in \emph{2022 IEEE International Joint
  Conference on Biometrics (IJCB)}.\hskip 1em plus 0.5em minus 0.4em\relax
  IEEE, 2022, pp. 1--10.

\bibitem{sarridis2023towards}
I.~Sarridis, C.~Koutlis, S.~Papadopoulos, and C.~Diou, ``Towards fair face
  verification: An in-depth analysis of demographic biases,'' in \emph{Joint
  European Conference on Machine Learning and Knowledge Discovery in
  Databases}.\hskip 1em plus 0.5em minus 0.4em\relax Springer, 2023, pp.
  194--208.

\bibitem{albiero2022face}
V.~Albiero, K.~W. Bowyer, and M.~C. King, ``Face regions impact recognition
  accuracy differently across demographics,'' in \emph{2022 IEEE International
  Joint Conference on Biometrics (IJCB)}.\hskip 1em plus 0.5em minus
  0.4em\relax IEEE, 2022, pp. 1--9.

\bibitem{bhatta2023gender}
A.~Bhatta, V.~Albiero, K.~W. Bowyer, and M.~C. King, ``The gender gap in face
  recognition accuracy is a hairy problem,'' in \emph{Proceedings of the
  IEEE/CVF Winter Conference on Applications of Computer Vision}, 2023, pp.
  303--312.

\bibitem{pangelinan2024exploring}
G.~Pangelinan, K.~Krishnapriya, V.~Albiero, G.~Bezold, K.~Zhang, K.~Vangara,
  M.~C. King, and K.~W. Bowyer, ``Exploring causes of demographic variations in
  face recognition accuracy,'' in \emph{Computer Vision}.\hskip 1em plus 0.5em
  minus 0.4em\relax Chapman and Hall/CRC, 2024, pp. 61--81.

\bibitem{ozturk2024can}
K.~Ozturk, H.~Wu, and K.~W. Bowyer, ``Can the accuracy bias by facial hairstyle
  be reduced through balancing the training data?'' in \emph{Proceedings of the
  IEEE/CVF Conference on Computer Vision and Pattern Recognition}, 2024, pp.
  1519--1528.

\bibitem{wu2024facial}
H.~Wu, S.~Tian, A.~Bhatta, K.~{\"O}zt{\"u}rk, K.~Ricanek, and K.~W. Bowyer,
  ``Facial hair area in face recognition across demographics: Small size, big
  effect,'' in \emph{Proceedings of the IEEE/CVF Winter Conference on
  Applications of Computer Vision}, 2024, pp. 1131--1140.

\bibitem{mamede2024fairness}
R.~M. Mamede, P.~C. Neto, and A.~F. Sequeira, ``Fairness under cover:
  Evaluating the impact of occlusions on demographic bias in facial
  recognition,'' \emph{arXiv preprint arXiv:2408.10175}, 2024.

\bibitem{kurz2025illusion}
P.~J. Kurz, H.~Wu, K.~W. Bowyer, and P.~Terh{\"o}rst, ``On the ``illusion'' of
  gender bias in face recognition: Explaining the fairness issue through
  non-demographic attributes,'' \emph{arXiv preprint arXiv:2501.12020}, 2025.

\bibitem{pangelinan2024analyzing}
G.~Pangelinan, A.~Bhatta, H.~Wu, M.~C. King, and K.~W. Bowyer, ``Analyzing the
  impact of demographic and operational variables on 1-to-many face id
  search,'' \emph{IEEE Transactions on Technology and Society}, 2024.

\bibitem{dantcheva2012can}
A.~Dantcheva, C.~Chen, and A.~Ross, ``Can facial cosmetics affect the matching
  accuracy of face recognition systems?'' in \emph{2012 IEEE Fifth
  International Conference on Biometrics: Theory, Applications and Systems
  (BTAS)}.\hskip 1em plus 0.5em minus 0.4em\relax IEEE, 2012, pp. 391--398.

\bibitem{kotwal2019detection}
K.~Kotwal, Z.~Mostaani, and S.~Marcel, ``Detection of age-induced makeup
  attacks on face recognition systems using multi-layer deep features,''
  \emph{IEEE Transactions on Biometrics, Behavior, and Identity Science},
  vol.~2, no.~1, pp. 15--25, 2019.

\bibitem{Liu_2022_CVPR}
C.~Liu, X.~Yu, Y.-H. Tsai, M.~Faraki, R.~Moslemi, M.~Chandraker, and Y.~Fu,
  ``Learning to learn across diverse data biases in deep face recognition,'' in
  \emph{Proceedings of the IEEE/CVF Conference on Computer Vision and Pattern
  Recognition (CVPR)}, June 2022, pp. 4072--4082.

\bibitem{zeng2021survey}
D.~Zeng, R.~Veldhuis, and L.~Spreeuwers, ``A survey of face recognition
  techniques under occlusion,'' \emph{IET Biometrics}, vol.~10, no.~6, pp.
  581--606, 2021.

\bibitem{boutros2021mfr}
F.~Boutros, N.~Damer, J.~N. Kolf, K.~Raja, F.~Kirchbuchner, R.~Ramachandra,
  A.~Kuijper, P.~Fang, C.~Zhang, F.~Wang \emph{et~al.}, ``Mfr 2021: Masked face
  recognition competition,'' in \emph{2021 IEEE International Joint Conference
  on Biometrics (IJCB)}.\hskip 1em plus 0.5em minus 0.4em\relax IEEE, 2021, pp.
  1--10.

\bibitem{kotwal2024latent}
K.~Kotwal, T.~Deshmukh, and P.~Gopal, ``Latent enhancing autoencoder for
  occluded image classification,'' in \emph{2024 IEEE International Conference
  on Image Processing (ICIP)}.\hskip 1em plus 0.5em minus 0.4em\relax IEEE,
  2024, pp. 894--900.

\bibitem{kortylewski2021compositional}
A.~Kortylewski, Q.~Liu, A.~Wang, Y.~Sun, and A.~Yuille, ``Compositional
  convolutional neural networks: A robust and interpretable model for object
  recognition under occlusion,'' \emph{International Journal of Computer
  Vision}, vol. 129, pp. 736--760, 2021.

\bibitem{albiero2021gendered}
V.~Albiero, K.~Zhang, M.~C. King, and K.~W. Bowyer, ``Gendered differences in
  face recognition accuracy explained by hairstyles, makeup, and facial
  morphology,'' \emph{IEEE Transactions on Information Forensics and Security},
  vol.~17, pp. 127--137, 2021.

\bibitem{bingham2017morph}
G.~Bingham, B.~Yip, M.~Ferguson, and C.~Nansalo, ``{MORPH-II: Inconsistencies
  and Cleaning},'' \emph{University of North Carolina Wilmington NSF REU},
  2017.

\bibitem{founds2011nist}
A.~P. Founds, N.~Orlans, W.~Genevieve, and C.~I. Watson, ``Nist special
  database 32-multiple encounter dataset ii (meds-ii),'' 2011.

\bibitem{eidinger2014age}
E.~Eidinger, R.~Enbar, and T.~Hassner, ``Age and gender estimation of
  unfiltered faces,'' \emph{IEEE Transactions on information forensics and
  security}, vol.~9, no.~12, pp. 2170--2179, 2014.

\bibitem{kai_curation_method}
V.~A. Kai~Zhang and K.~W. Bowyer, ``A method for curation of web-scraped face
  image datasets,'' in \emph{International Workshop on Biometrics and Forensics
  (IWBF)}, 2020.

\bibitem{xiong2018asian}
Z.~Xiong, Z.~Wang, C.~Du, R.~Zhu, J.~Xiao, and T.~Lu, ``An asian face dataset
  and how race influences face recognition,'' in \emph{Pacific Rim Conference
  on Multimedia}.\hskip 1em plus 0.5em minus 0.4em\relax Springer, 2018, pp.
  372--383.

\bibitem{cao2018vggface2}
Q.~Cao, L.~Shen, W.~Xie, O.~Parkhi, and A.~Zisserman, ``Vggface2: A dataset for
  recognising faces across pose and age,'' in \emph{Proceedings of the IEEE
  International Conference on Automatic Face \& Gesture Recognition}.\hskip 1em
  plus 0.5em minus 0.4em\relax IEEE, 2018, pp. 67--74.

\bibitem{hupont2019demogpairs}
I.~Hupont and C.~Fern{\'a}ndez, ``Demogpairs: Quantifying the impact of
  demographic imbalance in deep face recognition,'' in \emph{2019 14th IEEE
  international conference on automatic face \& gesture recognition (FG
  2019)}.\hskip 1em plus 0.5em minus 0.4em\relax IEEE, 2019, pp. 1--7.

\bibitem{georgopoulos2020investigating}
M.~Georgopoulos, Y.~Panagakis, and M.~Pantic, ``Investigating bias in deep face
  analysis: The kanface dataset and empirical study,'' \emph{Image and vision
  computing}, vol. 102, p. 103954, 2020.

\bibitem{wang2020mitigating}
M.~Wang and W.~Deng, ``Mitigating bias in face recognition using skewness-aware
  reinforcement learning,'' in \emph{Proceedings of the IEEE/CVF Conference on
  Computer Vision and Pattern Recognition}, 2020, pp. 9322--9331.

\bibitem{morales2020sensitivenets}
A.~Morales, J.~Fierrez, R.~Vera-Rodriguez, and R.~Tolosana, ``Sensitivenets:
  Learning agnostic representations with application to face images,''
  \emph{IEEE Transactions on Pattern Analysis and Machine Intelligence},
  vol.~43, no.~6, pp. 2158--2164, 2020.

\bibitem{robinson2023balancing}
J.~P. Robinson, C.~Qin, Y.~Henon, S.~Timoner, and Y.~Fu, ``Balancing biases and
  preserving privacy on balanced faces in the wild,'' \emph{IEEE Transactions
  on Image Processing}, 2023.

\bibitem{robinson2020face}
J.~P. Robinson, G.~Livitz, Y.~Henon, C.~Qin, Y.~Fu, and S.~Timoner, ``Face
  recognition: too bias, or not too bias?'' in \emph{Proceedings of the
  IEEE/CVF Conference on Computer Vision and Pattern Recognition Workshops},
  2020, pp. 0--1.

\bibitem{muhammad2021casia}
J.~Muhammad, Y.~Wang, C.~Wang, K.~Zhang, and Z.~Sun, ``Casia-face-africa: A
  large-scale african face image database,'' \emph{IEEE Transactions on
  Information Forensics and Security}, vol.~16, pp. 3634--3646, 2021.

\bibitem{guo2016ms}
Y.~Guo, L.~Zhang, Y.~Hu, X.~He, and J.~Gao, ``Ms-celeb-1m: A dataset and
  benchmark for large-scale face recognition,'' in \emph{Proceedings of 14th
  European Conference on Computer Vision (ECCV)}.\hskip 1em plus 0.5em minus
  0.4em\relax Springer, 2016, pp. 87--102.

\bibitem{kemelmacher2016megaface}
I.~Kemelmacher-Shlizerman, S.~M. Seitz, D.~Miller, and E.~Brossard, ``The
  megaface benchmark: 1 million faces for recognition at scale,'' in
  \emph{Proceedings of the IEEE Conference on Computer Vision and Pattern
  Recognition}, 2016, pp. 4873--4882.

\bibitem{fu2014learning}
S.~Fu, H.~He, and Z.-G. Hou, ``Learning race from face: A survey,'' \emph{IEEE
  Transactions on Pattern Analysis and Machine Intelligence}, vol.~36, no.~12,
  pp. 2483--2509, 2014.

\bibitem{levi2015age}
G.~Levi and T.~Hassner, ``Age and gender classification using convolutional
  neural networks,'' in \emph{Proceedings of the IEEE Conference on Computer
  Vision and Pattern Recognition Workshops}, 2015, pp. 34--42.

\bibitem{o2012demographic}
A.~J. O'Toole, P.~J. Phillips, X.~An, and J.~Dunlop, ``Demographic effects on
  estimates of automatic face recognition performance,'' \emph{Image and Vision
  Computing}, vol.~30, no.~3, pp. 169--176, 2012.

\bibitem{grother2022face}
P.~Grother, ``Face recognition vendor test (frvt) part 8: Summarizing
  demographic differentials,'' \emph{National Institute of Standards and
  Technology (NIST)}, vol. 8429, 2022.

\bibitem{conti2024assessing}
J.-R. Conti and S.~Cl{\'e}men{\c{c}}on, ``Assessing uncertainty in similarity
  scoring: Performance \& fairness in face recognition,'' in \emph{The Twelfth
  International Conference on Learning Representations}, 2024.

\bibitem{conti2022mitigating}
J.-R. Conti, N.~Noiry, S.~Clemencon, V.~Despiegel, and S.~Gentric, ``Mitigating
  gender bias in face recognition using the von mises-fisher mixture model,''
  in \emph{International Conference on Machine Learning}.\hskip 1em plus 0.5em
  minus 0.4em\relax PMLR, 2022, pp. 4344--4369.

\bibitem{de2021fairness}
T.~de~Freitas~Pereira and S.~Marcel, ``Fairness in biometrics: a figure of
  merit to assess biometric verification systems,'' \emph{IEEE Transactions on
  Biometrics, Behavior, and Identity Science}, vol.~4, no.~1, pp. 19--29, 2021.

\bibitem{schuckers2022statistical}
M.~Schuckers, S.~Purnapatra, K.~Fatima, D.~Hou, and S.~Schuckers, ``Statistical
  methods for assessing differences in false non-match rates across demographic
  groups,'' in \emph{International Conference on Pattern Recognition}.\hskip
  1em plus 0.5em minus 0.4em\relax Springer, 2022, pp. 570--581.

\bibitem{kotwal2022fairness}
K.~Kotwal and S.~Marcel, ``{Fairness Index Measures to Evaluate Bias in
  Biometric Recognition},'' in \emph{Proceedings of the International
  Conference on Pattern Recognition Workshops}.\hskip 1em plus 0.5em minus
  0.4em\relax Springer, 2022, pp. 479--493.

\bibitem{howard2022evaluating}
J.~J. Howard, E.~J. Laird, R.~E. Rubin, Y.~B. Sirotin, J.~L. Tipton, and A.~R.
  Vemury, ``Evaluating proposed fairness models for face recognition
  algorithms,'' in \emph{International Conference on Pattern Recognition
  Workshops}.\hskip 1em plus 0.5em minus 0.4em\relax Springer, 2022, pp.
  431--447.

\bibitem{villalobos2022fair}
E.~Villalobos, D.~Mery, and K.~Bowyer, ``Fair face verification by using
  non-sensitive soft-biometric attributes,'' \emph{IEEE Access}, vol.~10, pp.
  30\,168--30\,179, 2022.

\bibitem{elobaid2024sum}
A.~Elobaid, N.~Ramoly, L.~Younes, S.~Papadopoulos, E.~Ntoutsi, and
  I.~Kompatsiaris, ``Sum of group error differences: A critical examination of
  bias evaluation in biometric verification and a dual-metric measure,'' in
  \emph{2024 IEEE 18th International Conference on Automatic Face and Gesture
  Recognition (FG)}.\hskip 1em plus 0.5em minus 0.4em\relax IEEE, 2024, pp.
  1--9.

\bibitem{solano2024comprehensive}
I.~Solano, A.~Pe{\~n}a, A.~Morales, J.~Fierrez, R.~Tolosana,
  F.~Zamora-Martinez, and J.~S. Agustin, ``Comprehensive equity index (cei):
  Definition and application to bias evaluation in biometrics,'' in
  \emph{International Conference on Pattern Recognition}.\hskip 1em plus 0.5em
  minus 0.4em\relax Springer, 2024, pp. 110--126.

\bibitem{salvador2021faircal}
T.~Salvador, S.~Cairns, V.~Voleti, N.~Marshall, and A.~Oberman, ``Faircal:
  Fairness calibration for face verification,'' in \emph{International
  Conference on Learning Representations}, 2022.

\bibitem{terhorst2021comprehensive}
P.~Terh{\"o}rst, J.~N. Kolf, M.~Huber, F.~Kirchbuchner, N.~Damer, A.~M. Moreno,
  J.~Fierrez, and A.~Kuijper, ``A comprehensive study on face recognition
  biases beyond demographics,'' \emph{IEEE Transactions on Technology and
  Society}, vol.~3, no.~1, pp. 16--30, 2021.

\bibitem{gong2020jointly}
S.~Gong, X.~Liu, and A.~Jain, ``Jointly de-biasing face recognition and
  demographic attribute estimation,'' in \emph{Computer Vision--ECCV 2020: 16th
  European Conference, Glasgow, UK, August 23--28, 2020, Proceedings, Part XXIX
  16}.\hskip 1em plus 0.5em minus 0.4em\relax Springer, 2020, pp. 330--347.

\bibitem{terhorst2020comparison}
P.~Terh{\"o}rst, M.~L. Tran, N.~Damer, F.~Kirchbuchner, and A.~Kuijper,
  ``Comparison-level mitigation of ethnic bias in face recognition,'' in
  \emph{Proceedings of the International Workshop on Biometrics and
  Forensics}.\hskip 1em plus 0.5em minus 0.4em\relax IEEE, 2020, pp. 1--6.

\bibitem{kotwal2024WACV}
K.~Kotwal and S.~Marcel, ``Mitigating demographic bias in face recognition via
  regularized score calibration,'' in \emph{Proceedings of the IEEE/CVF Winter
  Conference on Applications of Computer Vision (WACV) Workshops}, January
  2024, pp. 1150--1159.

\bibitem{melzi2024frcsyn}
P.~Melzi, R.~Tolosana, R.~Vera-Rodriguez, M.~Kim, C.~Rathgeb, X.~Liu,
  I.~DeAndres-Tame, A.~Morales, J.~Fierrez, J.~Ortega-Garcia \emph{et~al.},
  ``{FRCSyn-onGoing: Benchmarking and comprehensive evaluation of real and
  synthetic data to improve face recognition systems},'' \emph{Information
  Fusion}, vol. 107, p. 102322, 2024.

\bibitem{melzi2024frcsyn1}
------, ``{FRCSyn challenge at WACV 2024: Face recognition challenge in the era
  of synthetic data},'' in \emph{Proceedings of the IEEE/CVF Winter Conference
  on Applications of Computer Vision}, 2024, pp. 892--901.

\bibitem{deandres2024frcsyn}
I.~DeAndres-Tame, R.~Tolosana, P.~Melzi, R.~Vera-Rodriguez, M.~Kim, C.~Rathgeb,
  X.~Liu, A.~Morales, J.~Fierrez, J.~Ortega-Garcia \emph{et~al.}, ``Frcsyn
  challenge at cvpr 2024: Face recognition challenge in the era of synthetic
  data,'' in \emph{Proceedings of the IEEE/CVF Conference on Computer Vision
  and Pattern Recognition}, 2024, pp. 3173--3183.

\bibitem{singh2022anatomizing}
R.~Singh, P.~Majumdar, S.~Mittal, and M.~Vatsa, ``Anatomizing bias in facial
  analysis,'' in \emph{Proceedings of the AAAI Conference on Artificial
  Intelligence}, vol.~36, 2022, pp. 12\,351--12\,358.

\bibitem{hort2024bias}
M.~Hort, Z.~Chen, J.~M. Zhang, M.~Harman, and F.~Sarro, ``Bias mitigation for
  machine learning classifiers: A comprehensive survey,'' \emph{ACM Journal on
  Responsible Computing}, vol.~1, no.~2, pp. 1--52, 2024.

\bibitem{deb2018longitudinal}
D.~Deb, N.~Nain, and A.~K. Jain, ``Longitudinal study of child face
  recognition,'' in \emph{2018 International Conference on Biometrics
  (ICB)}.\hskip 1em plus 0.5em minus 0.4em\relax IEEE, 2018, pp. 225--232.

\bibitem{schroff2015facenet}
F.~Schroff, D.~Kalenichenko, and J.~Philbin, ``Facenet: A unified embedding for
  face recognition and clustering,'' in \emph{Proceedings of the IEEE
  conference on computer vision and pattern recognition}, 2015, pp. 815--823.

\bibitem{kortylewski2019analyzing}
A.~Kortylewski, B.~Egger, A.~Schneider, T.~Gerig, A.~Morel-Forster, and
  T.~Vetter, ``Analyzing and reducing the damage of dataset bias to face
  recognition with synthetic data,'' in \emph{Proceedings of the IEEE/CVF
  Conference on Computer Vision and Pattern Recognition Workshops}, 2019, pp.
  0--0.

\bibitem{yucer2020exploring}
S.~Yucer, S.~Ak{\c{c}}ay, N.~Al-Moubayed, and T.~P. Breckon, ``Exploring racial
  bias within face recognition via per-subject adversarially-enabled data
  augmentation,'' in \emph{Proceedings of the IEEE/CVF Conference on Computer
  Vision and Pattern Recognition Workshops}, 2020, pp. 18--19.

\bibitem{kotwal2024demographic}
K.~Kotwal and S.~Marcel, ``Demographic fairness transformer for bias mitigation
  in face recognition,'' in \emph{2024 IEEE International Joint Conference on
  Biometrics (IJCB)}.\hskip 1em plus 0.5em minus 0.4em\relax IEEE, 2024, pp.
  1--10.

\bibitem{amini2019uncovering}
A.~Amini, A.~P. Soleimany, W.~Schwarting, S.~N. Bhatia, and D.~Rus,
  ``Uncovering and mitigating algorithmic bias through learned latent
  structure,'' in \emph{Proceedings of the 2019 AAAI/ACM Conference on AI,
  Ethics, and Society}, 2019, pp. 289--295.

\bibitem{wang2019deep}
P.~Wang, F.~Su, Z.~Zhao, Y.~Guo, Y.~Zhao, and B.~Zhuang, ``Deep class-skewed
  learning for face recognition,'' \emph{Neurocomputing}, vol. 363, pp. 35--45,
  2019.

\bibitem{yin2019feature}
X.~Yin, X.~Yu, K.~Sohn, X.~Liu, and M.~Chandraker, ``Feature transfer learning
  for face recognition with under-represented data,'' in \emph{Proceedings of
  the IEEE/CVF Conference on Computer Vision and Pattern Recognition}, 2019,
  pp. 5704--5713.

\bibitem{gong2021mitigating}
S.~Gong, X.~Liu, and A.~Jain, ``Mitigating face recognition bias via group
  adaptive classifier,'' in \emph{Proceedings of the IEEE/CVF Conference on
  Computer Vision and Pattern Recognition}, 2021, pp. 3414--3424.

\bibitem{huang2023gradient}
L.~Huang, M.~Wang, J.~Liang, W.~Deng, H.~Shi, D.~Wen, Y.~Zhang, and J.~Zhao,
  ``Gradient attention balance network: Mitigating face recognition racial bias
  via gradient attention,'' in \emph{Proceedings of the IEEE/CVF Conference on
  Computer Vision and Pattern Recognition}, 2023, pp. 38--47.

\bibitem{Ma_2023_ICCV}
J.~Ma, Z.~Yue, K.~Tomoyuki, S.~Tomoki, K.~Jayashree, S.~Pranata, and H.~Zhang,
  ``Invariant feature regularization for fair face recognition,'' in
  \emph{Proceedings of the IEEE/CVF International Conference on Computer Vision
  (ICCV)}, October 2023, pp. 20\,861--20\,870.

\bibitem{alasadi2019toward}
J.~Alasadi, A.~Al~Hilli, and V.~K. Singh, ``Toward fairness in face matching
  algorithms,'' in \emph{Proceedings of the 1st International Workshop on
  Fairness, Accountability, and Transparency in MultiMedia}, 2019, pp. 19--25.

\bibitem{liang2019additive}
J.~Liang, Y.~Cao, C.~Zhang, S.~Chang, K.~Bai, and Z.~Xu, ``Additive adversarial
  learning for unbiased authentication,'' in \emph{Proceedings of the IEEE/CVF
  Conference on Computer Vision and Pattern Recognition}, 2019, pp.
  11\,428--11\,437.

\bibitem{li2021learning}
Y.~Li, Y.~Sun, Z.~Cui, S.~Shan, and J.~Yang, ``Learning fair face
  representation with progressive cross transformer,'' \emph{arXiv preprint
  arXiv:2108.04983}, 2021.

\bibitem{xu2021consistent}
X.~Xu, Y.~Huang, P.~Shen, S.~Li, J.~Li, F.~Huang, Y.~Li, and Z.~Cui,
  ``Consistent instance false positive improves fairness in face recognition,''
  in \emph{Proceedings of the IEEE/CVF conference on computer vision and
  pattern recognition}, 2021, pp. 578--586.

\bibitem{wang2021meta}
M.~Wang, Y.~Zhang, and W.~Deng, ``Meta balanced network for fair face
  recognition,'' \emph{IEEE Transactions on Pattern Analysis and Machine
  Intelligence}, vol.~44, no.~11, pp. 8433--8448, 2021.

\bibitem{park2022fair}
S.~Park, J.~Lee, P.~Lee, S.~Hwang, D.~Kim, and H.~Byun, ``Fair contrastive
  learning for facial attribute classification,'' in \emph{Proceedings of the
  IEEE/CVF Conference on Computer Vision and Pattern Recognition}, 2022, pp.
  10\,389--10\,398.

\bibitem{serna2022sensitive}
I.~Serna, A.~Morales, J.~Fierrez, and N.~Obradovich, ``Sensitive loss:
  Improving accuracy and fairness of face representations with
  discrimination-aware deep learning,'' \emph{Artificial Intelligence}, vol.
  305, p. 103682, 2022.

\bibitem{zhang2022fairness}
F.~Zhang, K.~Kuang, L.~Chen, Y.~Liu, C.~Wu, and J.~Xiao, ``Fairness-aware
  contrastive learning with partially annotated sensitive attributes,'' in
  \emph{Proceedings of the International Conference on Learning
  Representations}, 2023.

\bibitem{wang2023mixfairface}
F.-E. Wang, C.-Y. Wang, M.~Sun, and S.-H. Lai, ``Mixfairface: Towards ultimate
  fairness via mixfair adapter in face recognition,'' in \emph{Proceedings of
  the AAAI Conference on Artificial Intelligence}, vol.~37, no.~12, 2023, pp.
  14\,531--14\,538.

\bibitem{li2025instance}
Y.~Li, Y.~Sun, Z.~Cui, P.~Shen, and S.~Shan, ``Instance-consistent fair face
  recognition,'' \emph{IEEE Transactions on Pattern Analysis and Machine
  Intelligence}, 2025.

\bibitem{michalski2018impact}
D.~Michalski, S.~Y. Yiu, and C.~Malec, ``The impact of age and threshold
  variation on facial recognition algorithm performance using images of
  children,'' in \emph{2018 International Conference on Biometrics
  (ICB)}.\hskip 1em plus 0.5em minus 0.4em\relax IEEE, 2018, pp. 217--224.

\bibitem{srinivas2019face}
N.~Srinivas, K.~Ricanek, D.~Michalski, D.~S. Bolme, and M.~King, ``Face
  recognition algorithm bias: Performance differences on images of children and
  adults,'' in \emph{Proceedings of the IEEE/CVF Conference on Computer Vision
  and Pattern Recognition Workshops}, 2019, pp. 0--0.

\bibitem{terhorst2020post}
P.~Terh{\"o}rst, J.~N. Kolf, N.~Damer, F.~Kirchbuchner, and A.~Kuijper,
  ``Post-comparison mitigation of demographic bias in face recognition using
  fair score normalization,'' \emph{Pattern Recognition Letters}, vol. 140, pp.
  332--338, 2020.

\bibitem{dhar2021pass}
P.~Dhar, J.~Gleason, A.~Roy, C.~D. Castillo, and R.~Chellappa, ``Pass:
  protected attribute suppression system for mitigating bias in face
  recognition,'' in \emph{Proceedings of the IEEE/CVF International Conference
  on Computer Vision}, 2021, pp. 15\,087--15\,096.

\bibitem{liu2022oneface}
J.~Liu, Z.~Yu, H.~Qin, Y.~Wu, D.~Liang, G.~Zhao, and K.~Xu, ``Oneface: one
  threshold for all,'' in \emph{European Conference on Computer Vision}.\hskip
  1em plus 0.5em minus 0.4em\relax Springer, 2022, pp. 545--561.

\bibitem{linghu2024ijcb}
Y.~Linghu, T.~de~Freitas~Pereira, C.~Ecabert, S.~Marcel, and M.~Günther,
  ``Score normalization for demographic fairness in face recognition,'' in
  \emph{2024 IEEE International Joint Conference on Biometrics (IJCB)}, 2024,
  pp. 1--11.

\bibitem{conti2024mitigating}
J.-R. Conti and S.~Cl{\'e}men{\c{c}}on, ``Mitigating bias in facial recognition
  systems: Centroid fairness loss optimization,'' in \emph{International
  Conference on Pattern Recognition}.\hskip 1em plus 0.5em minus 0.4em\relax
  Springer, 2024, pp. 371--385.

\bibitem{dhar2020towards}
P.~Dhar, J.~Gleason, H.~Souri, C.~D. Castillo, and R.~Chellappa, ``Towards
  gender-neutral face descriptors for mitigating bias in face recognition,''
  \emph{arXiv preprint arXiv:2006.07845}, 2020.

\bibitem{liu2021rectifying}
B.~Liu, S.~Zhang, G.~Song, H.~You, and Y.~Liu, ``Rectifying the data bias in
  knowledge distillation,'' in \emph{Proceedings of the IEEE/CVF International
  Conference on Computer Vision}, 2021, pp. 1477--1486.

\bibitem{paganini2020prune}
M.~Paganini, ``Prune responsibly,'' \emph{arXiv preprint arXiv:2009.09936},
  2020.

\bibitem{iofinova2023bias}
E.~Iofinova, A.~Peste, and D.~Alistarh, ``Bias in pruned vision models:
  In-depth analysis and countermeasures,'' in \emph{Proceedings of the IEEE/CVF
  Conference on Computer Vision and Pattern Recognition}, 2023, pp.
  24\,364--24\,373.

\bibitem{lin2022fairgrape}
X.~Lin, S.~Kim, and J.~Joo, ``Fairgrape: Fairness-aware gradient pruning method
  for face attribute classification,'' in \emph{European Conference on Computer
  Vision}.\hskip 1em plus 0.5em minus 0.4em\relax Springer, 2022, pp. 414--432.

\bibitem{caldeira2024mst}
E.~Caldeira, J.~S. Cardoso, A.~F. Sequeira, and P.~C. Neto, ``Mst-kd: Multiple
  specialized teachers knowledge distillation for fair face recognition,''
  \emph{arXiv preprint arXiv:2408.16563}, 2024.

\bibitem{jacob2018quantization}
B.~Jacob, S.~Kligys, B.~Chen, M.~Zhu, M.~Tang, A.~Howard, H.~Adam, and
  D.~Kalenichenko, ``Quantization and training of neural networks for efficient
  integer-arithmetic-only inference,'' in \emph{Proceedings of the IEEE
  conference on computer vision and pattern recognition}, 2018, pp. 2704--2713.

\bibitem{yucer2022does}
S.~Yucer, M.~Poyser, N.~Al~Moubayed, and T.~P. Breckon, ``Does lossy image
  compression affect racial bias within face recognition?'' in \emph{2022 IEEE
  International Joint Conference on Biometrics (IJCB)}.\hskip 1em plus 0.5em
  minus 0.4em\relax IEEE, 2022, pp. 1--10.

\bibitem{qiu2024gone}
T.~Qiu, A.~Nichani, R.~Tadayon, and H.~Jeong, ``Gone with the bits:
  Benchmarking bias in facial phenotype degradation under low-rate neural
  compression,'' in \emph{ICML 2024 Next Generation of AI Safety Workshop},
  2024.

\bibitem{atzori2023demographic}
A.~Atzori, G.~Fenu, and M.~Marras, ``Demographic bias in low-resolution deep
  face recognition in the wild,'' \emph{IEEE Journal of Selected Topics in
  Signal Processing}, vol.~17, no.~3, pp. 599--611, 2023.

\bibitem{george2024digi2real}
A.~George and S.~Marcel, ``Digi2real: Bridging the realism gap in synthetic
  data face recognition via foundation models,'' in \emph{Proceedings of the
  Winter Conference on Applications of Computer Vision Workshops}, 2025, pp.
  1469--1478.

\bibitem{huber2024bias}
M.~Huber, A.~T. Luu, F.~Boutros, A.~Kuijper, and N.~Damer, ``Bias and diversity
  in synthetic-based face recognition,'' in \emph{Proceedings of the IEEE/CVF
  Winter Conference on Applications of Computer Vision}, 2024, pp. 6215--6226.

\bibitem{fatima2024large}
K.~Fatima, M.~Schuckers, G.~Cruz-Ortiz, D.~Hou, S.~Purnapatra, T.~Andrews,
  A.~Neupane, B.~Marshall, and S.~Schuckers, ``A large-scale study of
  performance and equity of commercial remote identity verification
  technologies across demographics,'' in \emph{2024 IEEE International Joint
  Conference on Biometrics (IJCB)}.\hskip 1em plus 0.5em minus 0.4em\relax
  IEEE, 2024, pp. 1--8.

\end{thebibliography}



\end{document}